%% file: TUPI.tex
\documentclass[10pt,twocolumn,letterpaper]{article}

\usepackage{iccv}
\usepackage{times}
\usepackage{epsfig}
\usepackage{graphicx}
\usepackage{amsmath}
\usepackage{amssymb}

\usepackage{microtype}
\usepackage{graphicx}
\usepackage{subfigure}
\usepackage{booktabs}
\usepackage[usenames,dvipsnames]{xcolor}
\usepackage{amsmath}
\usepackage{amssymb}
\usepackage{multirow}
\usepackage{enumitem}
\usepackage{bbm}
\usepackage{algorithm}
\usepackage{algorithmic}

\newcommand{\IGNORE}[1]{}
\include{kimkimacros}

\usepackage[accsupp]{axessibility}  % Improves PDF readability for those with disabilities.

\usepackage[pagebackref=true,breaklinks=true,colorlinks,bookmarks=false]{hyperref}

\iccvfinalcopy 
\def\iccvPaperID{8308} % *** Enter the ICCV Paper ID here

\begin{document}

%%%%%%%%% TITLE
\title{Testing using Privileged Information \\
by Adapting Features with Statistical Dependence}

\author{Kwang In Kim\\
UNIST
\and
James Tompkin\\
Brown University
}

\maketitle

%%%%%%%%%%%%%%%%%%%%%%%%%%%%%%%%%%%%%%%%%%%%%%%%%%%%%%%%%%
%%%%%%%%% ABSTRACT
\begin{abstract}
Given an imperfect predictor, we exploit additional features \emph{at test time} to improve the predictions made, without retraining and without knowledge of the prediction function.
This scenario arises if training labels or data are proprietary, restricted, or no longer available, or if training itself is prohibitively expensive.
We assume that the additional features are useful if they exhibit strong statistical dependence to the underlying perfect predictor. 
Then, we empirically estimate and strengthen the statistical dependence between the initial noisy predictor and the additional features via manifold denoising.
As an example, we show that this approach leads to improvement in real-world visual attribute ranking.
\\
Project webpage: \href{http://www.jamestompkin.com/tupi}{http://www.jamestompkin.com/tupi}
\end{abstract}

%%%%%%%%%%%%%%%%%%%%%%%%%%%%%%%%%%%%%%%%%%%%%%%%%%%%%%%%%%
%%%%%%%%% BODY TEXT
\section{Introduction}

In supervised learning, sometimes we have additional information \emph{at training time} but not at testing time. One example is human action recognition, where training images have additional skeletal features like bone lengths~\cite{ShiKim17}. Another class of examples is tasks with crowdsourced labels, which have derived confidence values from multiple noisy human labelings~\cite{LapHeiSch14}. In training, each input and output pair $(\mbg,y)$ has additional `privileged' features $\mbh$ to help learn mapping function $f$. Vapnik and Vashist proposed to exploit these data with \emph{learning using privileged information}~\cite{VapIzm15}. Their support vector machine (SVM) training algorithm was later generalized to other learning algorithms.

We consider the complementary scenario where additional features $\mbh$ are only available \emph{at testing time}. We call this testing using privileged information (TUPI). 
This is different from retraining with new features: 
Given a pre-trained $f$ and test data $\mbg$ with \emph{test time} features $\mbh$, TUPI does not assume access to the original large set of labeled training data $\{(\mbg_i,y_i)\}$. The TUPI scenario allows for  the test-time privileged information to be heterogeneous, of unknown form, and not even necessarily useful to improving performance.

TUPI scenarios are more common than we might think. Predictors are now provided as closed-source libraries or cloud services by large technology companies, but smaller companies might have additional domain- or problem-specific information to integrate. As the initial predictor is proprietary with inaccessible training data, this integration is difficult. Another example is when a predictor was trained on a single lens camera but applied to multi-lens cameras (across smartphone models or robot build variations), say, if new depth signal data were available. TUPI is also useful when training requires specialized hardware that is unavailable at test time, or when data are so large or transient that storing them in perpetuity is difficult. 
%(Please see additional discussion in Sec.~\ref{sec:conclusion}.)

%%%%%%%%%%%%%%%%%%%%%%%%%%%%%%%%%%%%%%%%%%%%%%%%%%%%%%%%%%
\paragraph{Problem description.}
Suppose we have an estimation problem with an input feature space $\calG$ and the corresponding output space $\calY\subset \R$. Traditional algorithms aim to identify the underlying function $f^*:\calG\to\calY$ based on the input training features $G^{tr}=\{\mbg^{tr}_1,\ldots,\mbg^{tr}_l\}\subset \calG$ and the corresponding task-specific labels $Y^{tr}\subset \calY$. Once an estimate $f^I$ of $f^*$ is constructed, we can apply it to unseen test data points $G=\{\mbg_{1},\ldots,\mbg_n\}$ to construct the prediction $\mbf^I:=f^I|_{G}=[f^I(\mbg_{1}),\ldots,f^I(\mbg_n)]^\top$.

In the TUPI scenario, we assume that additional privileged information feature sets $\{H^i\}_{i=1}^m$ are provided for the test set $G$ such that each test instance $\mbg_{k}\in G$ is accompanied by $m$ additional features $\{\mbh^1_{k},\ldots,\mbh^m_{k}\}$. Our goal is to exploit $\{H^i\}$ to improve prediction $\mbf^I$. 
However, each new feature set $H^i$ may or may not be related to the underlying function $f^*$ and so may or may not be useful to improve $\mbf^I$. 
Further, we do not assume explicit forms for the feature extractors $g$ or $h^i$ (by which $G$ and $H^i$ are obtained), or for the learned function $f$. For example, $f$ could be a deep neural network (DNN) regressor or a rule-based classifier.
Lastly, we do not assume access to a large set of labeled data points $(G^{tr},Y^{tr})$ at the testing stage.
Otherwise, the problem could be solved by adding $\{H^i\}_{i=1}^m$ to the feature set and applying traditional supervised feature selection algorithms~\cite{SonSmoGre12}. In general, exploiting additional privileged information during testing to improve a prediction is an ill-posed problem, and existing algorithms are not able to accomplish this without having access to the data or labels, or are only applicable to test time features of a specific form.

%%%%%%%%%%%%%%%%%%%%%%%%%%%%%%%%%%%%%%%%%%%%%%%%%%%%%%%%%%
\paragraph{Denoising statistical dependence.}
Our insight is that if the test time feature sets are useful, then they will exhibit strong \emph{statistical dependence} with the underlying perfect predictor $f^*$. For instance, in the extreme case of known perfect statistical dependence, knowing $H^i$ would immediately identify $f^*$. Since we do not know in advance which feature sets (if any) are actually useful, we must estimate these dependencies. We regard the initial predictor $f^I$ as a noisy version of $f^*$, and we empirically estimate the pairwise dependencies between $f^I$ and each feature set $H^i$. Then, we selectively strengthen---or \emph{denoise}---the pairwise statistical dependencies to improve $f^I$. That is, we:

\begin{enumerate}[itemsep=0pt, parsep=2pt, topsep=1pt, partopsep=2pt, leftmargin=12pt]
\item Embed $f^I$ and $\{H^i\}_{i=1}^m$ into a model manifold $M$ where the Hilbert-Schmidt independence criterion, a consistent measure of statistical dependence, constitutes a similarity measure~\cite{GreBouSmo05}.
\item Strengthen dependencies via denoising on $M$~\cite{HeiMai07}.
\end{enumerate}

We demonstrate the TUPI scenario with visual attribute ranking~\cite{ParGra11}, in which users provide pair-wise rank comparisons for attribute-based database ordering. This problem is a good fit for TUPI because ranking applications commonly have multiple related feature representations and attributes. For this, we simply use the corresponding rank loss (Eq.~\ref{e:rankloss}). Through experiments across seven real-world datasets, we show that our approach can lead to significantly improved estimation accuracy over the initial predictors.

%%%%%%%%%%%%%%%%%%%%%%%%%%%%%%%%%%%%%%%%%%%%%%%%%%%%%%%%%%
\paragraph{Semi-supervised adaptation.}
Given the initial prediction $\mbf^I$ and the test time features $\{H^i\}$, our algorithm improves $\mbf^I$ in an unsupervised manner. However, like many unsupervised approaches, our method has (two) hyperparameters that must be tuned for best-possible performance. This issue has no good approach in the unsupervised setting. In practical applications, we must rely on users to sample and evaluate hyperparameters assuming that their selections would be guided by experience on related problems.

For our method, user sampling is feasible as our method induces smoothness in accuracy over hyperparameters and is fast to execute ($\approx$1 second on 50K items; see supplemental Fig.~\ref{f:hyperparameters}). This setting makes objective assessment challenging as the results are subject to user experience. Therefore, for comparison to other techniques, we use validation labels to tune these parameters. This renders the validation scenario of our adaptation algorithm semi-supervised. In our experiments, we further show that using such a small set of validation labels to retrain a new predictor is not competitive.

%%%%%%%%%%%%%%%%%%%%%%%%%%%%%%%%%%%%%%%%%%%%%%%%%%%%%%%%%%
\paragraph{Related work.}
As TUPI uses small validation label sets for hyperparameter optimization, it might bring to mind semi-supervised learning and one might consider building a graph Laplacian from the test time features to help solve the task~\cite{ZhoWesGre04}. In our experiments, we demonstrate that TUPI provides a stronger alternative to this simple baseline. TUPI might also bring to mind multi-task learning (MTL), where features or functions learned for one problem are applicable to other problems~\cite{EvgPon04,LeeYanHwa16}. The closest related work in this setting is Kim~\etal's predictor combination algorithm~\cite{KimTomRic17} that regards predictions from potentially-related tasks as test time features. However, this algorithm cannot be applied to arbitrary multi-dimensional test time features (see Sec.~\ref{s:experiments}). 

The most closely-related work is the CoConut framework by Khamis and Lampert~\cite{KhaLam15}. This regularizes the output label space during \emph{co-classification} to prefer certain class proportions, without knowing anything about the label predictors, by enforcing smoothness via a graph Laplacian measured over neighboring points. These points can be defined by original features or by test time features. The latter case leads to a scenario similar to TUPI, and so we adapt CoConut to our relative attributes setting. We find that the two approaches are broadly complementary: CoConut exploits the \emph{spatial smoothness} manifested via local neighborhood structure (\ie via a graph Laplacian) while our algorithm exploits \emph{statistical dependence} that is measured across the whole dataset. We show this explicitly in experiments that combine both algorithms (supplemental Fig.~\ref{f:TUPICoConutComb}).

%%%%%%%%%%%%%%%%%%%%%%%%%%%%%%%%%%%%%%%%%%%%%%%%%%%%%%%%%%
\section{Background}
\label{s:background}

\paragraph{Hilbert-Schmidt independence criterion (HSIC).} 
HSIC is a consistent measure of statistical \mbox{(in-)dependence} based on the reproducing kernel Hilbert space (RKHS) embeddings of probability distributions~\cite{GreBouSmo05}. Suppose we have two data spaces $\calV$ and $\calW$, equipped with joint probability distribution $\mbbP_{\mbv\mbw}$ and marginals $\mbbP_\mbv$ and $\mbbP_\mbw$, and we wish to estimate their statistical dependence. Unlike other popular dependence measures like mutual information, HSIC does not require estimating the underlying joint probability density. This is valuable because, in the TUPI scenario, this density may not even exist (see Sec.~\ref{s:technicalsection}). 

For $\calV$, we define a separable RKHS of functions $\calK_\mbv$ characterized by the feature map $\phi:\calV\to\calK_\mbv$ and the positive definite kernel function $k_\mbv(\mbv,\mbv'):=\langle\phi(\mbv),\phi(\mbv')\rangle$. 
Similarly, for $\calW$, we define $\calK_\mbw$, the feature map $\psi$, and the corresponding kernel $k_\mbw$. The HSIC associated with $\calK_\mbv$, $\calK_\mbw$, and $\mbbP_{\mbv\mbw}$ is:
\begin{align}
\text{HSIC}(\calK_\mbv,&\calK_\mbw,\mbbP_{\mbv\mbw}) = \E_{\mbv\mbv'\mbw\mbw'}[k_\mbv(\mbv,\mbv')k_\mbw(\mbw,\mbw')]\nonumber\\
&+\E_{\mbv\mbv'}[k_\mbv(\mbv,\mbv')]\E_{\mbw\mbw'}[k_\mbw(\mbw,\mbw')]\nonumber\\
& -2\E_{\mbv\mbw}\left[\E_{\mbv'}[k_\mbv(\mbv,\mbv')]\E_{\mbw'}[k_\mbw(\mbw,\mbw')]\right].\nonumber
\end{align}

For bounded and \emph{universal}~\cite{Ste01} kernels $k_\mbv$ and $k_\mbw$, such as Gaussian kernels (Eq.~\ref{e:Gaussian}), HSIC is well-defined and is zero only when the two distributions $\mbbP_\mbv$ and $\mbbP_\mbw$ are independent: 
HSIC is the maximum mean discrepancy (MMD) between the joint probability measure $\mbbP_{\mbv\mbw}$ and the product of marginals $\mbbP_{\mbv}\mbbP_{\mbw}$ computed with the product kernel $k_{\mbv\mbw}=K_\mbv \otimes K_\mbw$~\cite{MuaFukSri17}:
\begin{align}
\text{HSIC}(\calK_\mbv,\calK_\mbw,\mbbP_{\mbv\mbw}) &= \text{MMD}^2(\mbbP_{\mbv\mbw},\mbbP_{\mbv}\mbbP_{\mbw})\nonumber\\
&=\|\mu_k[\mbbP_{\mbv\mbw}]-\mu_k[\mbbP_{\mbv}\mbbP_{\mbw}]\|^2_k,
\end{align}
where $\|\cdot\|_k$ is the RKHS norm of $\calK_k$ and $\mu_k[\mbbP]$ is the \emph{kernel mean embedding} of $\mbbP$ based on $k$~\cite{MuaFukSri17}. If the kernels $K_\mbx$ and $K_\mby$ are universal, the MMD becomes a proper distance measure of probability distributions (\ie, $\text{MMD}(\mbbP_A,\mbbP_B)=0$ only when $\mbbP_A$ and $\mbbP_B$ are identical), which applied to the distance between the joint and marginal distributions corresponds to the condition of independence.

In practice, we do not have access to the underlying probability distributions, but only to a sample $\{\mbv_i,\mbw_i\}_{i=1}^n$ drawn from $\mbbP_{\mbv\mbw}$. However, HSIC is applicable to sample-based estimates: empirical HSIC estimates have uniform convergence guarantees to the underlying true HSIC, which provides a reliable finite sample estimate. Thus, we construct a sample-based HSIC estimate~\cite{GreBouSmo05}:
\begin{align}
\widehat{\text{HSIC}} = \tr[\mbK_\mbv \mbC \mbK_\mbw \mbC],\nonumber
\end{align}
where $[\mbK_\mbv]_{ij}=k_\mbv(\mbv_i,\mbv_j)$, $[\mbK_\mbw]_{ij}=k_\mbw(\mbw_i,\mbw_j)$, and $\mbC=I-\frac{1}{n}\ones\ones^\top$ with $\ones=[1,\ldots,1]^\top$.

%%%%%%%%%%%%%%%%%%%%%%%%%%%%%%%%%%%%%%%%%%%%%%%%%%%%%%%%%%
\paragraph{Manifold denoising.}
This estimates the underlying manifold structure from noisy samples within the ambient space of a manifold~\cite{GonShaMed10, WanTu13, KimTomRic17, HeiMai07}. A set of data points $P=\{\mbp_0,\ldots,\mbp_m\}\subset \R^d$ are assumed to be noisy samples of `true' data from an underlying embedded sub-manifold $M$ of $\R^d$, \ie, $\mbp_i = \imath(\mbq_i)+\epsilon$ for $\mbq_i\in M$, with embedding $\imath:M\to\R^d$. Assuming i.i.d.~Gaussian noise $\epsilon$ in the ambient space $\R^d$, Hein and Maier's manifold denoising algorithm suppresses noise in $P$---it pushes $P$ towards $M$---without having access to $M$ directly~\cite{HeiMai07}. 

We begin by building the graph Laplacian matrix $\Delta$ from pairwise similarities of $P$ in $\R^d$: $\Delta = I-D^{-1}W$:
\begin{align}
\quad [W]_{ij} &=\kappa(\|\mbp_i-\mbp_j\|^2,\sigma_P^2)\nonumber\\ 
&:= \exp\left(-\frac{\|\mbp_i-\mbp_j\|^2}{\sigma_P^2}\right).
\label{e:gaussiankernel}
\end{align}
The scale hyperparameter $\sigma^2_P>0$ and the diagonal matrix $D$ perform the probabilistic normalization: $[D]_{ii}=\sum_j [W]_{ij}$. Using $\Delta$ as the generator of a diffusion process on $P$, the denoising algorithm iteratively improves the solution $P$ by simulating the diffusion process governed by the differential equation with diffusion coefficient $\gamma>0$.
\begin{align}
\frac{\partial P}{\partial t} &= -\gamma\Delta P
\Rightarrow P(t+1)-P(t) = -\gamma \Delta P(t).
\label{e:diffusionequation}
\end{align}

In our approach, we embed the predicted evaluations $\mbf^I$ and test time features $\{H^i\}$ as points on a manifold to facilitate denoising $\mbf^I$ (Sec.~\ref{s:technicalsection}). In this case, only $\mbf$ is evolved while $\{H^i\}$ is fixed throughout diffusion. With this goal, at each discrete time step $t$, we construct $P$ by placing $\mbf(t)$ as the zero-th element (\ie, $\mbp_0(t)=\mbf(t)$,  $\mbp_0(0)=\mbf^I$), with the remaining elements in $P$ corresponding to $\{H^i\}$, and formulate the denoising of a single point $\mbp_0$ given $P\setminus\{\mbp_0\}$ as minimizing the energy:
\begin{align}
\label{e:impliciteuler}
\calO(\mbp) = \|\mbp-\mbp_0(t)\|^2+\lambda\sum_{i=1}^{m} \alpha_i\|\mbp-\mbp_i\|^2,
\end{align}
where $\alpha_i=[W(t)]_{1i}\big/\sum_{j}[W(t)]_{1j}$ for $1\leq i\leq m$ with $\lambda>0$ being a hyperparameter. We obtain $\calO$ based on implicit Euler time-discretization of the diffusion equation (Eq.~\ref{e:diffusionequation}), which is stable for any $\gamma>0$. We fix $\gamma$ at 1. The diffusion process is nonlinear: The weight matrix $W$ and the corresponding Laplacian $\Delta$ evolve over $t$.

%%%%%%%%%%%%%%%%%%%%%%%%%%%%%%%%%%%%%%%%%%%%%%%
\section{Feature adaptation at testing time}
\label{s:technicalsection}
We are now prepared to estimate the statistical dependence between each test time feature set and the underlying function $f^*$, based on the sample evaluations $\{H^i\}$ and noisy $\mbf^I$. The estimated dependencies are noisy, too, as they are based on $\mbf^I$, and so we apply manifold denoising to suppress noise in both the estimated dependencies and in $\mbf^I$. To facilitate the HSIC-based denosing, we embed the feature extractors and the predictor function into a Riemannian manifold of covariance operators or kernels $\{k\}$.

\paragraph{Manifold of predictors and feature extractors.}
Suppose that $\calX$ is the space of input data instances from which features are extracted, \eg, $\calX$ is the space of pixel-valued images. 
Let us assume a feature extractor $g:\calX\to \calG$ from which we construct the estimated predictor $f$, plus an additional class of feature extractors $\{h^i: \calX\to\calH^i\}_{i=1}^m$. 
We assume that all feature extractors are measurable. 
While we do not have direct access to $\{g,f,h^1,\ldots,h^m\}$, we do have the empirical evaluations of $f$ and $\{h^i\}$ on a sample $X\subset \calX$: $\mbf^I=f|_{g(X)}$ and $h^i|_{X}=H^i$. 
Further, we assume that the original data instance space $\calX$ is equipped with a probability distribution $\mbbP_\mbx$ inducing the corresponding probability distributions $\mbbP_{g}$ and $\mbbP_{h^i}$ in $\calG$ and $\calH^i$, respectively. Then, $f$ is given as an estimate constructed from the labeled data points $(G^{tr},Y^{tr})$ sampled from the joint distribution $\mbbP_{\mbx\mby}$.

Adopting the HSIC framework, we introduce a reproducing kernel Hilbert space (RKHS) on each element of the feature and predictor (evaluation) class: $\{\calY,\calH^1,\ldots,\calH^m\}$: For $f(\calG)\subset \calY$, the RKHS $\calK_f$ is defined with kernel $k_f:\calY\times\calY\to\R$. Similarly, the RKHS $\calK_i$ is defined with kernel $k_i:\calH^i\times\calH^i\to\R$. Using the fact that an RKHS is uniquely identified by its kernel, we can define new RKHSs of functions directly on the input space $\calX$: A reproducing kernel $\overline{k}_f:\calX\times\calX\to\R$ is induced from $k_f$ by application of the feature map $g$ and the predictor $f$: $\overline{k}_f(\mbx,\mbx'):=k_f(f \circ g(\mbx),f\circ g(\mbx'))$ for $\mbx,\mbx'\in \calX$.
The positive definiteness of $\overline{k}_f$ is guaranteed by the positive definiteness of $k_f$. Similarly, the RKHS $\overline{\calK}_i$ on $\calX$ corresponding to $k_i$ is defined based on $\overline{k}_i(\mbx,\mbx'):=k_i(h^i(\mbx),h^i(\mbx'))$.
We use a Gaussian kernel $k_f$ with the width parameter $\sigma_f^2$:
\begin{align}
k_f(f(\mbx),f(\mbx')) = \kappa(\|f(\mbx)-f(\mbx')\|^2,\sigma_f^2).
\label{e:Gaussian}
\end{align}
At this point, our input space $\calX$ is equipped with a data generating distribution $\mbbP_\mbx$, itself connected to multiple RKHSs, each constructed by a feature extractor (or predictor) and the corresponding RKHS. We will use this structure to characterize all feature extractors and predictors based on their respective induced kernels defined on $\calX$, which enables us to compare them in a unified framework.

\paragraph{Manifold embedding.}
We embed predictor $f$ and feature extractors $\{h^i\}$ into a space $M$ of normalized kernels:
\begin{align}
f\to \widetilde{k}_f:=(\overline{k}_f-\mu_{\overline{k}_f})/(\|\overline{k}_f-\mu_{\overline{k}_f}\|_{\overline{k}_f}),
\label{e:normalization}
\end{align}
where $\mu_{\overline{k}_f}$ is the mean embedding of $\mbbP_\mbx$ based on the kernel $\overline{k}_f$~\cite{MuaFukSri17}, \ie, $\mu_{\overline{k}_f} = \E_\mbx [\overline{k}(\mbx,\cdot)]$, and $\|\overline{k}\|_{\overline{k}_f}=\E_{\mbx\mbx'}|k(\mbx,\mbx')|$, both of which are well-defined for bounded kernels $k_f$ including Gaussian kernels (Eq.~\ref{e:Gaussian}).
Similarly, $h^i$ is embedded into $M$ based on $\widetilde{k}_i$.

The space $M$ is a Hilbert submanifold of an ambient Hilbert space $\calM$ with the inner product:\footnote{$M$ can have a (semi)-Riemannian structure if we identify the local neighborhood $N(p)$ of each point $p\in M$ with its tangent space $T_p(M)$. As we will only use the ambient space distance $\calM$, an explicit construction of the metric in $M$ is unnecessary.}
\begin{align}
\langle \overline{k}_f, \overline{k}_i\rangle_{\calM}=\E_{\mbx\mbx'}[\overline{k}_f(\mbx,\mbx')\overline{k}_i(\mbx,\mbx')].
\label{e:innerproduct}
\end{align}
Our $M$ construction is motivated by two points.
First, that the inner product $\langle\overline{k}_f,\overline{k}_i\rangle_{\calM}$ between two centered kernels $\overline{k}_f-\mu_{\overline{k}_f}$ and $\overline{k}_i-\mu_{\overline{k}_i}$ is precisely the HSIC of the predictor $f$ and the feature extractor $h^i$ as random variables:
\begin{align}
\langle\overline{k}_f,\overline{k}_i\rangle_{\calM} &= \E_{\mbx\mbx'}\big[\left(\overline{k}_f(\mbx,\mbx')-\E_{\mbx''}[\overline{k}_f(\mbx,\mbx'')]\right)\cdot\nonumber\\
&\left(\overline{k}_i(\mbx,\mbx')-\E_{\mbx''}[\overline{k}_i(\mbx,\mbx'')]\right)\big]\nonumber\\
&=\E_{f f'h^i (h^i)'}\big[\left(k_f(f,f')-\E_{f''}[k_f(f,f'')]\right)\cdot\nonumber\\
&\left(k_{i}(h^i,(h^i)')-\E_{(h^i)''}[k_{i}(h^i,(h^i)'')]\right)\big],\nonumber
\end{align}
where $f=f\circ g(\mbx)$, $f'=f\circ g(\mbx')$, and $f''=f\circ g(\mbx'')$.

Second, by noting that the scale normalization in the embedding (Eq.~\ref{e:normalization}) is essential in our denoising application. In application scenarios like feature selection~\cite{SonSmoGre12} or clustering~\cite{SonSmoGre07}, HSIC is used without normalization. However, in our denoising scenario, the prediction variables $f$ are directly optimized based on how HSIC is influenced by the kernel evaluations $\overline{k}_f$. In this case, one could scale HSIC without influencing the resulting statistical dependencies. For instance, in the simple case of the standard dot-product kernel ($\overline{k}(\mbx,\mbx')=\mbx^\top\mbx'$), HSIC becomes the Frobenius norm of the standard cross-covariance matrix. This can be arbitrarily increased by multiplying $f$ with a positive constant, but a constant scaling should not influence any reasonable measure of statistical dependence. With the normalization in Eq.~\ref{e:normalization}, the inner product $\langle\widetilde{k}_f,\widetilde{k}_i\rangle_{\calM}$ captures the same dependence information between $f$ and $h^i$ as HSIC, but with reduced influence of the scales of $f$ and $h^i$'s.

The two random variables $f$ and $h^i$ have joint functional dependence on $\mbx$, and so their joint probability density may not exist. Using HSIC enables estimating the statistical dependence even in this case, as it is estimated entirely based on kernel evaluations. This is not directly possible for some other dependence measures, \eg, mutual information.

In practice, we have sample evaluations $\mbf^I=f|_{G}$ (of size $n$) and corresponding features $\{H^i\}$. From this, we obtain a finite dimensional manifold $\widehat{M}$ with point-embedding $\mbf\to \widetilde{\mbK}_\mbf:=\mbK_\mbf \mbC/\|\mbK_\mbf\mbC\|_F$ where $[\mbK_\mbf]_{kl}=k_f([\mbf]_k,[\mbf]_l)$ and $\|\cdot\|_F$ is the Frobenius norm.
The inner-product (Eq.~\ref{e:innerproduct}) on $\widehat{M}$ becomes the HSIC estimate $\widehat{\text{HSIC}}$.

\paragraph{Exploiting test time information by denoising statistical dependence.}
With the manifold structure $\widehat{M}$, we are now ready to apply the denoising algorithm (Eq.~\ref{e:impliciteuler}). $\widehat{M}$ is a sub-manifold of a matrix Hilbert space $\widehat{\calM}$ where the inner-product between two points $\mbK_A$ and $\mbK_B$ can be calculated as $\langle \mbK_A, \mbK_B\rangle_{\widehat{\calM}}=\tr[\mbK_A\mbK_B]$ (\cf Eq.~\ref{e:innerproduct}). 
Then, we iteratively minimize an energy functional, replacing the Euclidean distance in $\calO$ (Eq.~\ref{e:impliciteuler}) with the ambient metric (restricted to $\widehat{M}$)  $d^2_{\widehat{\calM}}(\widetilde{\mbK}_\mbf,\widetilde{\mbK}_i):=1-\langle\widetilde{\mbK}_\mbf,\widetilde{\mbK}_i\rangle_{\widehat{\calM}}$:
\begin{align}
\label{e:denoisingalgorithm}
\calO(\mbf) &= d_{\widehat{\calM}}^2(\widetilde{\mbK}_\mbf,\widetilde{\mbK}_\mbf(t))+ \lambda  \sum_{i=1}^m w^i(t) d_{\widehat{\calM}}^2(\widetilde{\mbK}_\mbf,\widetilde{\mbK}_i),\\
w^i(t) &=\frac{\kappa(d_{\widehat{\calM}}^2(\widetilde{\mbK}_\mbf(t),\widetilde{\mbK}_i),\sigma_w^2)}{\sum_{j=1}^m\kappa(d_{\widehat{\calM}}^2(\widetilde{\mbK}_\mbf(t),\widetilde{\mbK}_j),\sigma_w^2)}.
\label{e:wupdate}
\end{align}
Note that $\mbf$ denotes the variable to be optimized (based on $\widetilde{\mbK}_\mbf$, which is a function of $\mbf$) while $\widetilde{\mbK}_\mbf(t)$ represents the results obtained from the previous time step $t$. In general, when $k$ is non-linear (as for Gaussian kernels in Eq.~\ref{e:Gaussian}), the optimization problem is non-convex. We optimize $\calO$ via gradient descent with $\mbf(0)$ obtained as the initial prediction $\mbf^I$. The time and memory complexities of optimizing $\calO$ are $O(mn^3)$ and $O(mn^2)$, respectively.  
Algorithm~\ref{a:mainalg} summarizes this proposed TUPI process. With this form, we see that our approach does not need to know the underlying function $f$, nor its feature extractor $g$, nor the function underlying the test time information $h$. This fulfills the TUPI scenario.

\paragraph{Algorithm interpretation.}
We solve the minimization problem at iteration $t$ by trading the deviation from the solution of iteration $t-1$ (first term in Eq.~\ref{e:denoisingalgorithm}) with the statistical dependence of $f$ and the feature extractors weighted by $\{w^i(t)\}$ (in Eq.~\ref{e:denoisingalgorithm}; see~Eq.~\ref{e:impliciteuler}). 
Each weight $w^i(t)$ is an increasing function of the estimated dependence strength at $t$-th step, which disregards outliers. 
The uniformity of the weights is controlled by the hyperparameter $\sigma_w^2$:

\begin{itemize}[itemsep=1pt, parsep=2pt, topsep=1pt, partopsep=2pt, leftmargin=12pt]
    \item As $\sigma_w^2\to\infty$, all features contribute equally to the minimization, which might include outliers. 
    \item As $\sigma_w^2\to 0$, the single most relevant (statistically dependent) feature influences the construction, which might neglect other less relevant but still beneficial features.
\end{itemize}

Due to kernelization (Eq.~\ref{e:Gaussian}) and normalization ($\widetilde{\mbK}$) in point-embedding, our algorithm applies when the absolute scale of predictions is irrelevant, \eg, for ranking. For classification and regression, we would store the scales and means, normalize and denoise, then restore the scales and means.

\begin{algorithm}[tb]
   \caption{TUPI algorithm}
   \label{a:mainalg}
\begin{algorithmic}
   \STATE {\bfseries Input:} Initial predictor evaluations $\mbf^I$; class of test time features $\{H^i\}_{i=1}^m$; hyperparameters $\lambda$ and $\sigma^2_w$ (Eq.~\ref{e:denoisingalgorithm}); (maximum iteration number $T$; see~Sec.~\ref{s:experiments});\\
   \STATE {\bfseries Output:} Denoised evaluations {$\mbf^O$};\\
   \STATE {$t = 0$;} $\mbf(t)=\mbf^I$;\\
   \REPEAT
      \STATE Calculate weights $\{w^i(t)\}$ based on Eq.~\ref{e:wupdate};\\
      \STATE Update $\mbf(t)$ by minimizing $\calO$ (Eq.~\ref{e:denoisingalgorithm});\\
      \STATE $t$ = $t$+1;
   \UNTIL{termination condition is met (\eg if $t\geq T$);}
\end{algorithmic}
\end{algorithm}

\paragraph{Large-scale problems.}
When the time $O(mn^3)$ and memory $O(mn^2)$ complexities of optimizing $\calO$ are limiting, we adopt the Nystr\"{o}m approximation of $\mbK_\mbf$:
\begin{align}
\mbK_\mbf\approx \mbK_{\mbf B} \mbK_{BB}^{-1} \mbK_{\mbf B}^\top,
\label{e:kapprox}
\end{align}
where $[\mbK_{\mbf B}]_{kl}=k_f(b_k,b_l)$ for the basis set $B=\{b_1,\ldots,b_K\}$ and $[\mbK_{\mbf B}]_{kl} = k_f([\mbf]_k,b_l)$. The rank $K$ of the approximation is prescribed based on the computational and memory capacity limits. Similarly, each $\mbK_i$ is approximated based on the corresponding basis set ($\mbK_i\approx \mbK_{iB} [\mbK^i_{BB}]^{-1} \mbK_{iB}^\top$). 
For example, the second (unnormalized) trace term in Eq.~\ref{e:denoisingalgorithm} and its derivative with respect to $\mbf$ are written as:
\begin{align}
\label{e:HSICeval}
\tr[\mbK_\mbf &\mbC \mbK_i \mbC] \approx \calC(\mbf)= \tr[\mbK_{\mbf B} \mbS_{\mbf i}], \nonumber\\
\frac{\partial \calC(\mbf)}{\partial [\mbf]_k}&=2[\partial {\mbK_{\mbf B}}]_{(k,:)}[\mbS_{\mbf i}]_{(:,k)}, \nonumber\\
\mbS_{\mbf i}&=\mbK_{BB}^{-1} \mbK_{\mbf B}^\top \mbC \mbK_{iB}[\mbK^i_{BB}]^{-1} \mbK_{iB}^\top \mbC,
\end{align}
where $[A]_{(k,:)}$ denotes the $k$-th row of $A$ and $[\partial {\mbK_{\mbf B}}]_{kl}$ corresponds to the derivative of $k_f([\mbf]_k,b_l)$ (see Eq.~\ref{e:Gaussian}). 
The computational bottleneck in the gradient evaluation is the multiplication $\mbK_{\mbf B}^\top \mbC \mbK_{iB}$ for each $i=1,\ldots,m$, which takes $O(mnK^2)$ time. Thus, complexity is linear in the number of data points $n$ and number of test time features $m$.

\paragraph{Convergence.}
While the trajectory of the solution during iteration depends on the initial solution and the mean curvature of the manifold~\cite{HeiMai07}, in the limit case (as $t\to \infty$), the embedding $\widetilde{\mbK}_\mbf(t)$ of the solution $\mbf(t)$ becomes a weighted average of feature kernels; \ie, the solution becomes independent of the initial predictions $\widetilde{\mbK}_\mbf(0)$, which is not useful. This is analogous to conventional diffusion where, when all points are evolved, the solution converges towards a constant as $t\to\infty$ \cite{HeiMai07}. In our case, we evolve only the predictor embedding $\widetilde{\mbK}_\mbf(t)$ and hold the remaining test time feature embeddings $\{\widetilde{\mbK}_i\}$ fixed; hence, $\widetilde{\mbK}_\mbf(t)$ becomes the weighted average of $\{\widetilde{\mbK}_i\}$. Therefore, we must terminate the iteration before convergence (see Sec.~\ref{s:experiments}).

%%%%%%%%%%%%%%%%%%%%%%%%%%%%%%%%%%%%%%%%%%
\section{Experimental results}
\label{s:experiments}

\paragraph{Setting.}
We test our approach on the \emph{Relative Attributes} rank setting~\cite{ParGra11}. Our algorithm receives the initial rank evaluation $\mbf^I$ and the test time feature set $\{H^i\}$, and outputs an improved rank estimate $\mbf^O$. Throughout all experiments, the initialization $\mbf^I$ is predicted from 200 data points with pairwise comparison labels, by either a deep neural network (DNN) or a rank support vector machine (RSVM)~\cite{ChaKee10}---whichever gave the higher validation accuracy. Hyperparameters were optimized on the validation set: RSVM regularization parameter, DNN training epochs, DNN MLP layers (2--8) and neurons per layer (5--160). To optimize the DNN, we use a standard mini-batch gradient descent with batch normalization. For both DNN and RSVM, we use the soft hinge loss $l_H$: an ordered training pair $(q,r)$ implies that the ranking of $\mbg_q$ should be higher than $\mbg_r$:
\begin{align}
l_H((\mbg_q,\mbg_r);f)=\max\left(0,1-( f(\mbg_q)-f(\mbg_r) ) \right)^2.\label{e:rankloss}
\end{align}
For all datasets, we ran experiments 10 times with different training and validation sets and averaged the results. Accuracy is measured as the ratio of correct pairwise rank comparisons with respect to all possible pairs.

\paragraph{TUPI parameters and effect discussion.}
Our algorithm requires setting values for the large-scale factorization rank $K$ (Eq.~\ref{e:kapprox}), kernel scale  $\sigma^2_f$ (Eq.~\ref{e:Gaussian}), weight uniformity $\sigma_w^2$ (Eq.~\ref{e:denoisingalgorithm}), number of iterations $T$, and regularization $\lambda$.

We fix the approximation rank $K$ at 50 (Eq.~\ref{e:kapprox}). For multi-dimensional features, the basis points $\{b_k\}$ are constructed as their cluster centers. For one-dimensional features and for rank predictors $\mbf$, we obtain the basis points as the end-points of linearly sampled intervals in the respective ranges. While we expected the sample-based HSIC to increase in accuracy as $K$ increases, the performance for $K=200$ was not significantly higher than for $K=50$.

For kernel scale parameter $\sigma^2_f$, we use the standard heuristic and set it to twice the standard deviations of the pairwise distances of elements of $\mbf$. We set the scale parameter for each feature similarly.

We tune the remaining hyperparameters $\sigma_w^2$ and $\lambda$ based on 50 validation data points, with $T$ set implicitly: We set the maximum $T$ value at 50 and monitored the progress of validation accuracy: we terminate iteration immediately whenever the validation accuracy did not increase from the previous iteration. The influences of $\sigma_w^2$ and $T$ are complementary: Larger $\sigma_w^2$ and $T$ values set the focus on more strongly-dependent features, which leads to similar results as with smaller $\sigma_w^2$ and $T$ values.

\paragraph{Baselines.}
Since we are not aware of any existing algorithms for rank testing with test time information, we form comparisons with existing methods that apply to less general or alternative settings. We aim to show that na\"{i}vely applying existing algorithms to TUPI is challenging.

Our baselines are: 1) the initial rank prediction $\mbf^I$, for which all results are shown as \emph{relative difference} from this; 2) retraining a DNN or RSVM (whichever is better) on $\mbh$ with the validation labels (as we assume that a small validation set is available); 3) using semi-supervised learning (SSL) to build a graph Laplacian with the test time features and validation labels~\cite{ZhoWesGre04}, 4) Khamis and Lampert's CoConut algorithm~\cite{KhaLam15} adapted to TUPI in our rank prediction setting (in principle, their method is complementary to ours), and 5) Kim~\etal's predictor combination algorithm~\cite{KimTomRic17}. 
The hyperparameters of all baseline algorithms were tuned based on validation sets. 

\vspace{0.2cm}
\noindent\emph{Adapting CoConut~\cite{KhaLam15}.} We minimize the energy
\begin{align}
\label{e:orgcoconut}
\calO'(\mbv) = \|\mbv-\mbf^I\|^2+\frac{\lambda^C}{k^C} \mbv^\top L \mbv,
\end{align}
where $L$ is the graph Laplacian calculated based on local $k$-nearest neighbors (with $k=k^C$) in the test time feature space, and $\lambda^C$ and $k^C$ are hyperparameters. The first term in $\calO'$ ensures that the final solution does not deviate significantly from $\mbf^I$ while the second term contributes to improving the final solution by enforcing its spatial smoothness measured via the Laplacian $L$. Supplemental Section~\ref{s:coconut} provides the details of this adaptation and the construction of the Laplacian $L$.

\vspace{0.2cm}
\noindent\emph{Adapting Kim~\etal~\cite{KimTomRic17}.} This method forms predictive \emph{distributions} from reference tasks, then penalizes their pairwise KL-divergence from the target distribution. To adapt their method to our setting, if we let their reference task predictions be new features $\{H^i\}$, then this approach works when the (probability) space of each feature coincides with the space of predictions $\calY$. This makes their algorithm applicable only to datasets where one-dimensional test time features are provided as the predictions made for potentially-related tasks (\eg, the \emph{PubFig} and \emph{Shoes} datasets discussed shortly). We demonstrate that our general multi-dimensional test time feature algorithm is a strong alternative to Kim~\etal's approach even in this special setting.

%%%%%%%%%%%%%%%%%%%%%%%%%%%%%%%%%%%%%%%%%%%%%%%%%%%%%%%%%%%%%%%%%%%%%%%
\subsection{Results}
\paragraph{\emph{MFeat}.}
This contains 6 different feature sets (F1--F6) of 2,000 handwritten digits, with rank outputs obtained from digit class labels. We use each feature set as the baseline features $\mbg$, with the remaining features used as test time features $\{H^i\}_{i=1}^5$, creating 6 different experimental settings.

\begin{table}[t]
\caption{\emph{MFeat} dataset. Ranking algorithm mean accuracy percent, plus standard deviation in parenthesis, given the F1--F6 features.
$\mbf^I$: \hspace{0.71cm}The initial predictions.\newline
$\mbf^{O}$: \hspace{0.56cm} TUPI with other F-feature sets as test time information.\newline
$\mbf^{R}$: \hspace{0.66cm}$\mbf^{O}$ with an additional 10 random features.\newline
$\mbf^{S_1,S_3}$: \hspace{0.11cm} $\mbf^{O}$ with only randomly selected 1 and 3 F-feature sets.\newline
$\mbf^{G_1-G_3}$: $\mbf^{O}$ with ground-truth target variables as test time features (of decreasing noise standard deviations $\{1,0.2,0\}$).
}
\label{t:mfeatresults}
\vspace{0.25cm}
\resizebox{\linewidth}{!}
{
\begin{tabular}{l c c c c c c c c}
\toprule
& $\mbf^I$ & $\mbf^O$ & $\mbf^{R}$ & $\mbf^{S_1}$ & $\mbf^{S_3}$ & $\mbf^{G_1}$ & $\mbf^{G_2}$ & $\mbf^{G_3}$\\
\midrule
\multirow{2}{*}{F1} &77.85 &81.97 &82.14 &79.88 &82.15 &78.84 &87.84 &99.56 \\
&(2.26) &(2.95) &(2.86) &(1.49) &(2.37) &(2.35) &(0.84) &(0.34) \\
\midrule
\multirow{2}{*}{F2} &79.28 &81.45 &81.61 &80.40 &81.37 &80.00 &88.16 &99.61 \\
&(1.23) &(1.69) &(1.60) &(1.37) &(1.54) &(1.22) &(0.28) &(0.16) \\
\midrule
\multirow{2}{*}{F3} &75.70 &78.31 &78.25 &77.12 &78.32 &76.77 &87.13 &99.35 \\
&(2.38) &(3.13) &(3.16) &(1.79) &(3.02) &(2.20) &(0.82) &(0.42) \\
\midrule
\multirow{2}{*}{F4} &70.88 &74.33 &74.36 &70.80 &74.29 &72.00 &86.22 &99.53 \\
&(1.23) &(5.00) &(5.03) &(1.44) &(5.02) &(1.38) &(0.48) &(0.48) \\
\midrule
\multirow{2}{*}{F5} &76.05 &78.19 &78.05 &77.62 &78.06 &77.22 &87.35 &99.30 \\
&(2.66) &(3.29) &(3.26) &(3.28) &(3.72) &(2.62) &(0.72) &(0.44) \\
\midrule
\multirow{2}{*}{F6} &77.10 &82.25 &82.44 &79.30 &80.71 &78.07 &86.93 &99.44 \\
&(1.60) &(2.09) &(1.90) &(2.39) &(2.58) &(1.37) &(0.70) &(0.28) \\
\bottomrule
\end{tabular}
}
\end{table}

Our approach consistently improves performance ($\mbf^O$ in the second column of Table~\ref{t:mfeatresults}). TUPI utility is demonstrated by results for F1, F2, and F6, with higher accuracy than the highest individual feature (F2). Further, we verify our algorithm's ability to \emph{only pick useful features} by adding 10 additional random features to the test time information ($\{H^i\}_{i=1}^5$), with dimensions varying from 2 to 20. The results ($\mbf^{R}$) are similar to that of $\mbf^O$ without the spurious random features. In addition, we measured the sensitivity of our algorithm against the number of test time feature sets by randomly selecting only one and three feature sets out of five. The results (Table~\ref{t:mfeatresults}: $\mbf^{S_1,S_3}$) indicate that the performance of our algorithm gracefully degrades. 

Lastly, we verify the correct operation of TUPI in the ideal case by using ground-truth target ranks as test time information (Table~\ref{t:mfeatresults}: $\mbf^{G_1-G_3}$): The target rank variables are globally scaled to $[0,1]$ and contaminated with zero-mean Gaussian noise with standard deviations of $\{1,0.2,0\}$. This leads to average peak signal-to-noise ratios of $\{-0.13,14.08,\infty\}$-dB, respectively. When the noise-level is zero, our algorithm was able to fully exploit the test time information and achieve almost perfect rankings. Performance degrades gracefully as the noise-level increases.

\begin{figure*}
\centering
\includegraphics[width=\linewidth]{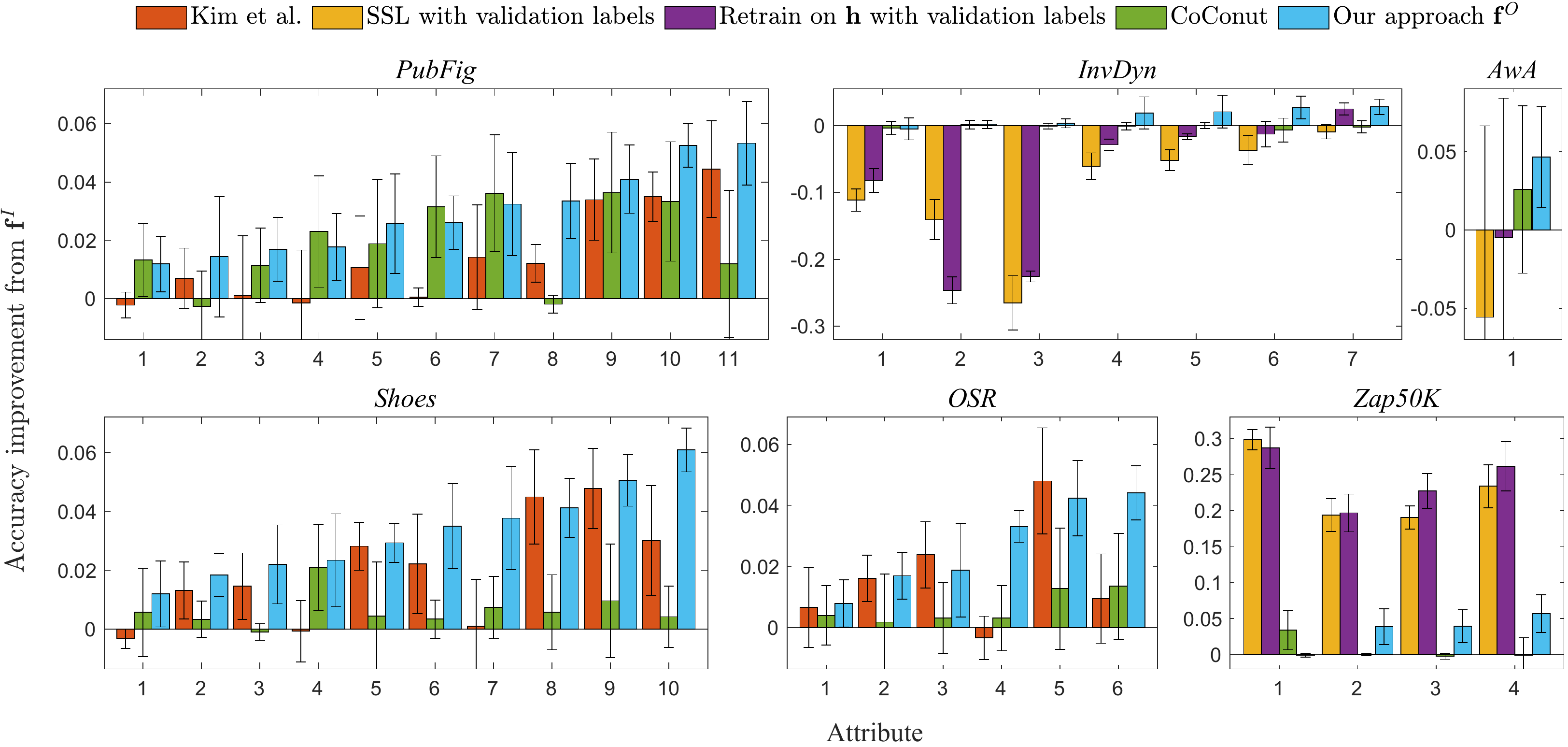}
\caption{Accuracy improvements over $\mbf^I$ for six datasets. Higher is better; error bars are one standard deviation above and below the mean. Kim~\etal~cannot apply to \emph{AwA}, \emph{Zap50K}, or \emph{InvDyn}. Attributes are sorted in ascending order of $\mbf^O$ accuracy values for improved readability. We include absolute accuracy values in supplemental Figure~\ref{f:resultsabsolute}. }
\label{f:resultsrelative}
\end{figure*}

\paragraph{\emph{PubFig}, \emph{Shoes}, and \emph{OSR}.} These contain 772 images of 8 classes with 11 attributes~\cite{ParGra11}, 14,658 images of 10 categories with 10 attributes~\cite{KovParGra12}, and 2,688 images of 6 attributes from 8 categories~\cite{KovParGra12}, respectively. The goal is to estimate rankings for each target attribute. The labels are provided as category-wise comparisons, \ie, each category has a stronger or weaker presence of certain attributes than other categories. For \emph{PubFig} and \emph{Shoes}, we construct the initial rankings $\mbf^I$ using the GIST features and color histograms provided by Parikh and Grauman~\cite{ParGra11}. Similarly, for \emph{OSR}, $\mbf^I$ was constructed using GIST features provided by the authors of \cite{KovParGra12}. We train a rank predictor from each target attribute. Then, for each attribute rank predictor, we use all other attribute rank predictions as test time information. 

This setting has been explored by Kim~\etal~\cite{KimTomRic17} as each test time feature set is only one-dimensional, though this method is not applicable for general test time features. For \emph{PubFig}, Kim~\etal's predictor combination algorithm largely improves performance over the baseline $\mbf^I$ (Fig.~\ref{f:resultsrelative}), with our algorithm ($\mbf^O$) making further improvements to attributes 7--11. CoConut also improves the performance from $\mbf^I$ and it generates the best results in attributes 1, 4, 6, and 7; in this respect CoConut and our algorithm are complementary. In the supplemental Section~\ref{s:coconuttupicombination}, we demonstrate that by combining these two approaches, we can obtain even better combination algorithms. For \emph{Shoes}, Kim~\etal's algorithm provides similar performance, with our algorithm further improving attributes 3, 4, 6, 7, and 10. Our algorithm constantly outperformed CoConut. For \emph{OSR}, our algorithm and Kim~\etal's algorithm are complementary: For attributes 3 and 5, Kim~\etal's algorithm is the best; for the remaining features, our algorithm further significantly improves the performance. CoConut achieves moderate accuracy gain. The accompanying supplemental Table~\ref{t:sigtestresults} provides tests of statistical significance of the results of all algorithms.

\paragraph{\emph{InvDyn}.} 
This contains 45,000 data points collected from robot movement tasks. Each point is a 21-dimensional feature vector containing 7 joint positions, velocities, and accelerations~\cite{VijSch00}. The goal is to estimate the 7 torque variables constituting 7-different ranking problems. For each target output, only a subset of 15 input measurements are provided as the baseline features $\mbg$, with the remaining measurements provided as test time features $\{H^i\}$. Using only the test time features $\mbh$ and SSL severely degrades the ranking performance overall. This indicates that the test time information is complementary to the baseline features, which our TUPI approach exploits to increase accuracy.

In our automatic evaluation, we use 50 data points to select the hyperparameters of our algorithm. Thus, for comparison, we also show the results of training a new ranker and the graph Laplacian-based semi-supervised learning (SSL) ranker on 50 data points (denoted as $\mbh$ and SSL in Fig.~\ref{f:resultsrelative}). Even with only 50 labeled data points, these rankers can show higher accuracies than the baseline $\mbf^I$, especially for attribute 6; however, they are considerably worse than $\mbf^I$ for attribute 1. It is rare for our approach to degrade performance, which helps demonstrate that our algorithm can exploit new features appropriately when a large label set is not available. CoConut did not show any noticeable improvement over $\mbf^I$.

\paragraph{Animals with Attributes dataset (\emph{AwA}).} This contains 30,475 images of 50 animals class. Our goal is to rank images according to class labels. We use the features extracted by a pre-trained DeCAF network~\cite{DonJiaVin14} as the baseline features $G$, and adopt SURF, PHOG, and VGG19 as the test time features $\{H^i\}_{i=1}^3$. These are provided by Lampert~\etal~\cite{LamNicHar09}. 

Again, CoConut improved upon the baseline rankers $\mbf^I$ while our approach further significantly improved performance. We observe that powerful VGG19 features provide more distinctive descriptions of features and lead to improvements: performance of TUPI using only VGG19 test time features is almost the same as using three test time features. Throughout the denoising process, our algorithm successfully \emph{selected} these VGG19 features.

\paragraph{Limitation---\emph{Zap50K} dataset.} 
When the test time information is significantly more powerful than the baseline features, retraining with even a small number of new labels can provide better results. This is demonstrated with the \text{\emph{Zap50K}} dataset, which contains 50,025 images of shoes with 4 attributes. Attribute labels are collected by instance-level pairwise comparison via \emph{Mechanical Turk}~\cite{YuGra14}.

We use the 30-dimensional color histogram features and 960-dimensional GIST features as $G$ and $H$, respectively, as provided by Yu and Grauman~\cite{YuGra14}. 
Across attributes, we use training and validation sets of around 300--400 pairs, and use test sets of around 300 pairs. $\mbf^I$ and $\mbh$ are trained on $G$ and $H$ with training and validation labels of the same size, respectively. In this case, GIST features $H$ lead to much higher accuracy than color histogram features $G$.
Even after our algorithm's improvements, a large performance gap remains---a potential upper limit of what our algorithm can achieved in TUPI on this dataset.

\section{Conclusion}
\label{sec:conclusion}

Testing using privileged information, or TUPI, considers how to improve a predictor by exploiting additional features that are only available during predictor testing. Our thesis is that these features can be useful if they exhibit strong statistical dependence to the underlying task predictor. This might not be true in general; supplemental Section~\ref{s:simplefailurecase} shows a toy limitation case where the additional features are identical to the initial predictions. However, in practice, we introduce a new algorithm that estimates and strengthens statistical dependence. Over seven real-world relative attribute ranking experiments, our algorithm improves performance over the baseline predictors (43/45 attributes) and, more importantly, only rarely degrades performance (2/45 attributes). This provide evidence that our thesis holds in practical applications, even when the feature adaptation scenario allows us no assumptions on the predictor or feature extractor forms, or any known existing statistical dependence.

Our experiments focused on two application scenarios using standard databases: 1) When predictors are trained on classical features and later tested with more powerful features; 2) when predictors are trained on a single feature but applied to multiple complementary features that are not necessary stronger than the original feature. Additional real-world application scenarios arise in the TUPI context:

\begin{enumerate}[itemsep=0pt, topsep=2pt, partopsep=2pt, leftmargin=12pt]
    \item When training requires hardware unavailable to practitioners, \eg, massive `tech giant'-scale cloud resources~\cite{HinVinDen14} or TPU, FPGA, or neuromorphic chips.
    \item When training data are `transient' and deleted after predictor creation, \eg, due to large storage requirements from scientific instruments like particle accelerators (CERN's petabyte-scale data~\cite{Anthony18}), but where new data are later available from new experiments to potentially exploit. A contrary scenario related to Vapnik and Vashist's framework~\cite{VapIzm15} would be when privileged data existed only when training the original predictor.
    \item Issues due to privacy concerns and data protection laws. For example, GDPR Right to be Forgotten/Erasure, where predictors trained on deleted data may still be kept, but where users re-using a service would provide new test time features (such as item ratings/recommendations).
\end{enumerate}

\paragraph{Performance bounds.}
Providing performance bounds to support our empirical findings requires new theoretical analysis techniques. The challenges are twofold: 1) Our algorithm builds upon statistical dependence via HSIC, rather than via a common probability distribution distance (\eg, KL-divergence). Even if we assume that the test time features contain ground-truth labels, the analysis of convergence towards the ground truth is not straightforward as most existing techniques are developed based on probability distribution distances (\eg, PAC Bayesian bounds). 2) We use $f^I$ as a surrogate to $f^*$ when estimating the statistical dependence between $f^*$ and $\{H^i\}_{i=1}^m$ and therefore, the deviations between $f^I$ and $f^*$ need to be quantified.

\section*{Acknowledgments}

KK was supported by a National Research Foundation of Korea (NRF) grant (No. 2021R1A2C2012195), Institute of Information and communications Technology Planning and evaluation (IITP) grant (2021--0--00537, Visual Common Sense Through Self-supervised Learning for Restoration of Invisible Parts in Images), and IITP grant (2020--0--01336, Artificial Intelligence Graduate School Program, UNIST), funded by the Korea government (MSIT).
This material is based on research sponsored by Defense Advanced Research Projects Agency (DARPA) and Air Force Research Laboratory (AFRL) under agreement number FA8750-19-2-1006. The U.S. Government is authorized to reproduce and distribute reprints for Governmental purposes notwithstanding any copyright notation thereon. The views and conclusions contained herein are those of the authors and should not be interpreted as necessarily representing the official policies or endorsements, either expressed or implied, of Defense Advanced Research Projects Agency (DARPA) and Air Force Research Laboratory (AFRL) or the U.S. Government.

%%%%%%%%%%%%%%%%%%%%%%%%%%%%%%%%%%%%%%%%%%%%%%%%%%%%%%%%%
%%%%%%%%%%%%%%%%%%%%%%%%%%%%%%%%%%%%%%%%%%%%%%%%%%%%%%%%%
{\small
\bibliographystyle{ieee_fullname}
\bibliography{./bib/biblio}
}

%%%%%%%%%%%%%%%%%%%%%%%%%%%%%%%%%%%%%%%%%%%%%%%%%%%%%%%%%
%%%%%%%%%%%%%%%%%%%%%%%%%%%%%%%%%%%%%%%%%%%%%%%%%%%%%%%%%

\clearpage
\appendix
\section{Supplemental material}

This supplemental material provides additional related work discussion (Sec.~\ref{s:additionaldiscussion}), additional background information to the Hilbert-Schmidt independence criterion used in our TUPI algorithm (Sec.~\ref{s:hsicbrief}) and in the visual attributes rank learning problem on which we evaluate our TUPI approach (Sec.~\ref{s:relativeattr}), and a brief summary of our denoising algorithm with an algorithm description (Sec.~\ref{s:algorithm}). We also provide further experimental details and results including tests of statistical significance, parameter sensitivity, and simple failure cases (Sec.~\ref{s:experiments}), plus additional future work ideas (Sec.~\ref{s:addfuturework}). Lastly, we present details of our adaptations of related baselines (Sec.~\ref{s:adaptation}): of Kim~\etal's algorithm~\cite{KimTomRic17} that only applies to one-dimensional test time features (Sec.~\ref{s:kimadaptation}); and of the CoConut algorithm proposed by Khamis and Lampert~\cite{KhaLam15} (Sec.~\ref{s:coconut}) for TUPI in visual attribute ranking problems. We also detail our approaches to combine our algorithm with CoConut to exploit these two complementary algorithms for increased performance (Sec.~\ref{s:coconuttupicombination}). Some contents from the main paper are reproduced so that this material is self-contained and easier to read.

%%%%%%%%%%%%%%%%%%%%%%%%%%%%%%%%%%%%%%%%%%%%%%%
\section{Additional related work details}
\label{s:additionaldiscussion}

In the main paper (Section 1), we discuss how the TUPI problem is related to---but different from---works in semi-supervised learning, in multi-task learning, and in predictor combination.

Domain adaptation is another related class of work to TUPI: Here, an estimator trained on a data domain $\calG$ equipped with a probability distribution $\mbbP_\mbg$ is tested on data generated from an updated probability distribution $\mbbP'_\mbg$ on the same domain $\calG$. This setting leads to new algorithms enabling adaptation of estimators in test-time using data sampled from $\mbbP'_\mbg$ as auxiliary information~\cite{RoyLam15}. However, these algorithms focus on modeling the change of distributions in the same domain. In contrast, in our TUPI setting, we use test time data sampled from multiple heterogeneous feature domains $\{\calH_i\}$ and thus our contribution is complementary to domain adaptation as a problem.

\subsection{Related HSIC applications} 
The Hilbert-Schmidt independence criterion (HSIC) has been successfully applied in clustering~\cite{SonSmoGre07} and domain adaptation~\cite{TuiCam16}. Particularly relevant works are Song~\etal's feature selection algorithm~\cite{SonSmoGre12}, which receives a set of features and task-specific labels as random variables and outputs the most statistically relevant features measured by HSIC, plus Gevaert~\etal's kernel learning framework that tunes the kernel parameters to maximize the dependence between the task labels and features (kernels)~\cite{GevPerVos16}. For regression, instead of maximizing the dependence between the prediction and features, Mooij~\etal minimize the dependence between the features and the residuals---the deviations between the estimated function values and the observed ground truths---to enable regression independently of the unknown noise generation process~\cite{MooJanPet09}. Our approach is inspired by these algorithms; however, as they are designed for use in training, they cannot be straightforwardly applied to the TUPI scenario without non-trivial adaptation.

%%%%%%%%%%%%%%%%%%%%%%%%%%%%%%%%%%%%%%%%%%%%%%%
\section{Additional background}

\subsection{The Hilbert-Schmidt independence criterion}
\label{s:hsicbrief}
Suppose we have two data spaces $\calV$ and $\calW$, equipped with joint probability distribution $\mbbP_{\mbv\mbw}$ and marginals $\mbbP_\mbv$ and $\mbbP_\mbw$, respectively. For $\calV$, we define a separable reproducing kernel Hilbert space (RKHS) $\calK_\mbv$ of functions characterized by the feature map $\phi:\calV\to\calK_\mbv$ and the positive definite kernel function $k_\mbv(\mbv,\mbv'):=\langle\phi(\mbv),\phi(\mbv')\rangle$.\footnote{A separable Hilbert space has a countable orthonormal basis facilitating the introduction of Hilbert-Schmidt operators.} 
The RHKS $\calK_\mbw$, and the corresponding kernel $k_\mbw$ and feature map $\psi$ are similarly defined for $\calW$. The cross-covariance operator associated with the joint probability distribution $\mbbP_{\mbv\mbw}$ is a linear operator $C_{\mbv\mbw}:\calK_\mbv\to \calK_\mbw$ which generalizes the cross-covariance matrix in Euclidean spaces:
\begin{align}
C_{\mbv\mbw}=\E_{\mbv\mbw}\left[(\phi(\mbv)-\E_{\mbv}\phi(\mbv))\otimes(\psi(\mbw)-\E_{\mbw}\psi(\mbw))\right],\nonumber
\end{align}
where $\otimes$ is the tensor product. Given this operator, the Hilbert-Schmidt independence criterion (HSIC) associated with $\calK_\mbv$, $\calK_\mbw$, and $\mbbP_{\mbv\mbw}$ is defined as the Hilbert-Schmidt norm of $C_{\mbv\mbw}$ which generalizes the Frobenius norm defined for matrices to operators~\cite{GreBouSmo05}:
\begin{align}
\text{HSIC}(\calK_\mbv,&\calK_\mbw,\mbbP_{\mbv\mbw}) = \|C_{\mbv\mbw}\|_{HS}^2\nonumber\\
&= \E_{\mbv\mbv'\mbw\mbw'}[k_\mbv(\mbv,\mbv')k_\mbw(\mbw,\mbw')]\nonumber\\
&+\E_{\mbv\mbv'}[k_\mbv(\mbv,\mbv')]\E_{\mbw,\mbw'}[k_\mbw(\mbw,\mbw')]\nonumber\\
&-2\E_{\mbv\mbw}\left[\E_{\mbv'}[k_\mbv(\mbv,\mbv')]\E_{\mbw'}[k_\mbw(\mbw,\mbw')]\right].
\end{align}
HSIC is defined as long as the kernels $k_\mbv$ and $k_\mbw$ are bounded, and it is always non-negative~\cite{GreBouSmo05}. Furthermore, when $k_\mbv$ and $k_\mbw$ are \emph{universal}~\cite{Ste01}, such as when they are Gaussian, HSIC is zero only when the two distributions $\mbbP_\mbv$ and $\mbbP_\mbw$ are independent: HSIC is the maximum mean discrepancy (MMD) between the joint probability measure $\mbbP_{\mbv\mbw}$ and the product of marginals $\mbbP_{\mbv}\mbbP_{\mbw}$ computed with the product kernel $k_{\mbv\mbw}=k_\mbv \otimes k_\mbw$~\cite{MuaFukSri17}:
\begin{align}
\text{HSIC}(\calK_\mbv,\calK_\mbw) &= \text{MMD}^2(\mbbP_{\mbv\mbw},\mbbP_{\mbv}\mbbP_{\mbw})\nonumber\\
&=\left\|\mu_k[\mbbP_{\mbv\mbw}]-\mu_k[\mbbP_{\mbv}\mbbP_{\mbw}]\right\|_k,\nonumber
\end{align}
where $\|\cdot\|_k$ is the RKHS norm of $\calK_k$ and $\mu_k[\mbbP]$ is the \emph{kernel mean embedding} of $\mbbP$ based on $k$~\cite{MuaFukSri17}. If the kernels $k_\mbv$ and $k_\mbw$ are universal, the MMD becomes a proper distance measure of probability distributions (\ie, $\text{MMD}(\mbbP_A,\mbbP_B)=0$ only when $\mbbP_A$ and $\mbbP_B$ are identical), which applied to the distance between the joint and marginal distributions corresponds to the condition of independence.

In practice, we do not have access to the underlying probability distributions but only a sample $\{\mbx_i,\mby_i\}_{i=1}^n$ drawn from $\mbbP_{\mbv\mbw}$. As such, we must construct a sample-based HSIC estimate:
\begin{align}
\widehat{\text{HSIC}} = \tr[\mbK_\mbv \mbC \mbK_\mbw \mbC],\nonumber
\end{align}
where $[\mbK_\mbv]_{ij}=k_\mbv(\mbv_i,\mbv_j)$, $[\mbK_\mbw]_{ij}=k_\mbw(\mbw_i,\mbw_j)$, and $\mbC=I-\frac{1}{n}\ones\ones^\top$ with $\ones=[1,\ldots,1]^\top$. The estimate $\widehat{\text{HSIC}}$ converges to the true HSIC with $O(1/\sqrt{n})$~\cite{GreBouSmo05}.

\subsection{The visual attribute rank learning problem}
\label{s:relativeattr}

Binary object labels describing the \emph{presence or absence} of attributes might be useful for automatic search~\cite{KovParGra12,YuGra15}. However, binary descriptors are insufficient for many attributes. Imagine shopping for shoes: there is no clear boundary between `sporty' and `non sporty' shoes. However, it is easy as a human to state that one shoe is `sportier' than another. Thus, the Parikh and Grauman approach of measuring \emph{relative attributes} (RA)~\cite{ParGra11} has broadened attribute-based data analysis to abstract and non-categorical labels.

To facilitate the training of automatic attribute predictors, users rank pairs of data points to describe their relationships: object $\mbx_i$ exhibits a stronger/weaker presence of attribute $A$ than $\mbx_j$. This technique can be thought of as implicitly introducing a global \emph{ranking function} to a dataset for a given attribute: There is a function $f^*$ such that $f^*(\mbx_i)>f^*(\mbx_j)$ implies that the rank of $\mbx_i$ is higher than that of $\mbx_j$.

Suppose we have a set of input features $G^{tr}=\{\mbg^{tr}_1,\ldots,\mbg^{tr}_l\}\subset \calG$ representing the underlying objects $X^{tr}=\{\mbx^{tr}_1,\ldots,\mbx^{tr}_l\}\subset \calX$ via a feature extractor $g:\calX\mapsto\calG$ ($g(\mbx^{tr}_i)=\mbg^{tr}_i$ for $1\leq i \leq l$) and the corresponding pairwise rank labels $R = \{(i(r),j(r))\}_r\subset \{1,\ldots,l\}\times \{1,\ldots,l\}$: $(i,j)\in R$ implies that the rank of $\mbx_{i}$ is higher than $\mbx_{j}$. An estimate $f$ of $f^*$ can be constructed by minimizing the average rank loss:
\begin{align}
L(f) &= \sum_{(i,j)\in R}l((\mbx_i,\mbx_j);f),\nonumber\\
l((\mbx_i,\mbx_j);f)&=\max\left(0,1-( f(\mbg_i)-f(\mbg_j) ) \right)^2.\nonumber
\end{align}
Once an estimate $f$ is constructed, we can apply it to unseen test data points $G=\{\mbg_{1},\ldots,\mbg_n\}$ to construct the prediction $\mbf^I:=f|_{G}=[f(\mbg_{1}),\ldots,f(\mbg_n)]^\top$.

%%%%%%%%%%%%%%%%%%%%%%%%%%%%%%%%%%%%%%%%%%%%%%%
\section{Summary of our denoising algorithm\\ for TUPI}
\label{s:algorithm}
Suppose we are given an initial predictor $\mbf(0)=\mbf^I$ and a set of test time features $\{H^i\}_{i=1}^m$. Our algorithm improves $\mbf(t)$ by embedding the predictor and the test time features into a manifold $\widehat{M}$ of (centered and scaled) kernel matrices, and performing manifold denoising therein:
\begin{align}
\mbf(t)&\to \widetilde{\mbK}_\mbf(t):=\frac{\mbK_\mbf(t)\mbC}{\|\mbC\mbK_\mbf(t)\mbC\|_F}\nonumber\\
H^i&\to \widetilde{\mbK}_i:=\frac{\mbK_i \mbC}{\|\mbC\mbK_i\mbC\|_F},\nonumber 
\end{align}
where $\|A\|_F$ is the Frobenius norm of $A$, and $\mbK_\mbf$ and $\mbK_i$ are the kernel matrices constructed from $\mbf$ and $H^i$, respectively:
\begin{align}
\label{e:kernelmat}
[\mbK_\mbf]_{kl}&=k_f([\mbf]_k,[\mbf]_l)= \exp\left(-\frac{\|[\mbf]_k-[\mbf]_l\|^2}{\sigma_f^2}\right)\\
[\mbK_i]_{kl}&=k_i(\mbh^i_k,\mbh^i_l)= \exp\left(-\frac{\|\mbh^i_k-\mbh^i_l\|^2}{\sigma_i^2}\right)\nonumber
\end{align}
with scale hyperparameters $\sigma_f^2$ and $\{\sigma_i^2\}_{i=1}^m$. Here, $\mbh^i_k$ denotes the $k$-th element of the feature set $H^i=\{\mbh^i_1,\ldots,\mbh^i_n\}$. $\{\sigma^2_f\}$ is set to be twice the standard deviations of pairwise distances of elements of $\mbf$; $\{\sigma_i^2\}_{i=1}^m$ are tuned similarly. 

This process is instantiated as iterative minimization of an energy functional $\calO(\cdot;t)$
\begin{align}
\label{e:denoisingalgorithmsup}
\calO(\mbf;t) &= d_{\widehat{\calM}}^2(\widetilde{\mbK}_\mbf,\widetilde{\mbK}_\mbf(t))\nonumber\\
&+ \lambda  \sum_{i=1}^m \left(\frac{w^i(t)}{\sum_{j=1}^m w^j(t)}\right) d_{\widehat{\calM}}^2(\widetilde{\mbK}_\mbf,\widetilde{\mbK}_i)\\
\label{e:weightupdate}
w^i(t) &=\exp\left(-\frac{d_{\widehat{\calM}}^2(\widetilde{\mbK}_\mbf(t),\widetilde{\mbK}_i)}{\sigma_w^2}\right),
\end{align}
where $d_{\widehat{\calM}}^2(\widetilde{\mbK}_A,\widetilde{\mbK}_B)=1-\tr[\widetilde{\mbK}_A\widetilde{\mbK}_B]$.

The number of iterations is a hyperparameter. In the experiments, we set the maximum iteration number $T$ at 50 and monitored the accuracy progress: we finish the iteration immediately whenever the validation accuracy did not increase from the previous iteration. Algorithm~\ref{a:mainalgsup} summarizes the TUPI denoising process.

% Forward declaration
\begin{figure*}[tb]
\centering
\includegraphics[width=\linewidth]{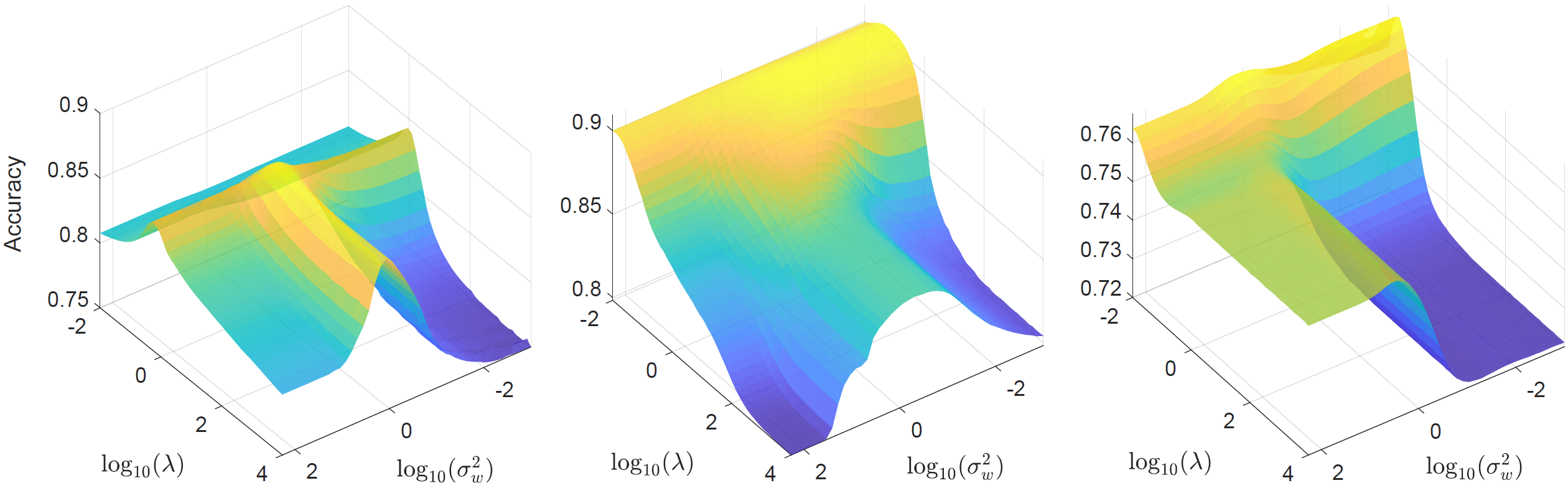}
\vspace{0.05cm}
\caption{Accuracy of our algorithm on \emph{PubFig} (left; attribute 2), \emph{OSR} (center; attribute 2), and \emph{Shoes} (right; attribute 5) datasets with respect to hyperparameters $\lambda$ (Eq.~\ref{e:denoisingalgorithmsup}) and $\sigma_w^2$ (Eq.~\ref{e:weightupdate}) varying over logarithmic intervals.}
\label{f:hyperparameters}
\end{figure*}

\begin{algorithm}[tb]
   \caption{TUPI algorithm}
   \label{a:mainalgsup}
\begin{algorithmic}
   \STATE {\bfseries Input:} Initial predictor evaluations $\mbf^I$; class of test time features $\{H^i\}_{i=1}^m$; hyperparameters $\lambda$ and $\sigma^2_w$ (Eq.~\ref{e:denoisingalgorithmsup}); (maximum iteration number $T$);\\
   \STATE {\bfseries Output:} Denoised evaluations {$\mbf^O$};\\
   \STATE {$t = 0$;}\\
   \STATE $\mbf(t)=\mbf^I$;\\
   \REPEAT
      \STATE Calculate weights $\{w^i(t)\}$ based on Eq.~\ref{e:weightupdate};\\
      \STATE Update $\mbf(t)$ by minimizing $\calO$ (Eq.~\ref{e:denoisingalgorithmsup});\\
      \STATE $t$ = $t$+1;
   \UNTIL{termination condition is met (\eg if $t\geq T$);}
\end{algorithmic}
\end{algorithm}

\subsection{Large scale problems}

For tasks where the time ($O(mn^3)$) and memory ($O(mn^2)$) complexities of optimizing $\calO$ are limiting, we adopt the Nystr\"{o}m approximation of kernel matrix $\mbK_\mbf$:\footnote{Readers are referred to \cite{ZhaFil17} for other sparse approximations including random Fourier features and block-averaged statistic.}
\begin{align}
\mbK_\mbf\approx \mbK_{\mbf B} \mbK_{BB}^{-1} \mbK_{\mbf B}^\top,
\label{e:kapproxsup}
\end{align}
where $[\mbK_{\mbf B}]_{kl}=k_f(b_k,b_l)$ for the basis set $B=\{b_1,\ldots,b_K\}$ and $[\mbK_{\mbf B}]_{kl} = k_f([\mbf]_k,b_l)$. The rank $K$ of the approximation is based on computational and memory capacity limits. Similarly, each $\mbK_i$ is approximated based on the corresponding basis set ($\mbK_i\approx \mbK_{iB} [\mbK^i_{BB}]^{-1} \mbK_{iB}^\top$). For example, the second (unnormalized) trace term in Eq.~\ref{e:denoisingalgorithmsup} and its derivative with respect to $\mbf$ are written as:
\begin{align}
\label{e:HSICevalsup}
\tr[\mbK_\mbf \mbC \mbK_i &\mbC] \approx \calC(\mbf)\nonumber\\
=& \tr[\mbK_{\mbf B} \mbK_{BB}^{-1} \mbK_{\mbf B}^\top \mbC \mbK_{iB}
[\mbK^i_{BB}]^{-1} \mbK_{iB}^\top \mbC]\\
\frac{\partial \calC(\mbf)}{\partial [\mbf]_k}=&2[\partial {\mbK_{\mbf B}}]_{(k,:)}\cdot\nonumber\\
&\Big[\mbK_{BB}^{-1} \mbK_{\mbf B}^\top \mbC \mbK_{iB}[\mbK^i_{BB}]^{-1} \mbK_{iB}^\top \mbC\Big]_{(:,k)},
\end{align}
where $[A]_{(k,:)}$ denotes the $k$-th row of $A$ and $[\partial {\mbK_{\mbf B}}]_{kl}$ corresponds to the derivative of $k_f([\mbf]_k,b_l)=\exp(-\|[\mbf]_k-b_l\|^2/\sigma_f^2)$. The main computational bottleneck in the gradient evaluation is the multiplication $\mbK_{\mbf B}^\top \mbC \mbK_{iB}$ for each $i=1,\ldots,m$, which takes total $O(m\times n\times K^2)$ time. As such, the complexity is linear in the number of data points $n$ and the number of test time features $m$.

\subsection{TUPI parameter smoothness}

As our algorithm is unsupervised, in practical applications, we expect users to evaluate different hyperparameter combinations. Figure~\ref{f:hyperparameters} shows that this approach is feasible as the rank accuracy surface with respect to the two parameters is smooth, enabling practical sampling approaches.

%%%%%%%%%%%%%%%%%%%%%%%%%%%%%%%%%%%%%%%%%%%%%%%
\section{Additional experimental details}
\label{s:experimentssup}

\subsection{Dataset details}
\paragraph{Multiple Features (\emph{MFeat}).} This dataset contains 6 different feature representations of 2,000 handwritten digits: Each input digit is represented by F1) 76 Fourier coefficients, F2) 216 profile correlations, F3) 64 Karhunen-Lo\`{e}ve coefficients, F4) 240 local color averages, F5) 47 Zernike moments, and F6) 6 morphological features~\cite{BlaMer98}. The target rank outputs are obtained based on digit class labels. In the main paper, we use each single feature set F1--F6 as the baseline features $\mbg$, and use the remaining features at test time $\{H^i\}_{i=1}^5$. 

\paragraph{Public Figure Faces (\emph{PubFig}),
\emph{Shoes}, and Outdoor Scene Recognition (\emph{OSR}) datasets.} \emph{PubFig} contains 772 images of 8 people with 11 attributes~\cite{ParGra11}. The goal is to estimate rankings on each of the target attributes: \emph{Masculine-looking}, \emph{White}, \emph{Young}, \emph{Smiling}, \emph{Chubby}, \emph{Visible-forehead}, \emph{Bushy-eyebrows}, \emph{Narrow-eyes}, \emph{Pointy-nose}, \emph{Big-lips}, \emph{Round-face}. The labels are provided as category-wise comparisons, \ie, each category (person for \emph{PubFig}) has a stronger or weaker presence of certain attributes than other categories. \emph{Shoes} dataset contains 14,658 images of 10 attributes and 10 categories~\cite{KovParGra12}: \emph{pointy-at-the-front}, \emph{open}, \emph{bright-in-color}, \emph{covered-with-ornaments}, \emph{shiny}, \emph{high-at-the-heel}, \emph{long-on-the-leg}, \emph{formal}, \emph{sporty}, and \emph{feminine}. \emph{OSR} contains 2,688 images of 6 attributes from 8 categories: \emph{natural}, \emph{open}, \emph{perspective}, \emph{large-objects}, \emph{diagonal-plane}, and \emph{close-depth}. Similar to \emph{PubFig}, the rank labels for \emph{Shoes} and \emph{OSR} are constructed from pairwise category-wise comparisons.

For \emph{PubFig} and \emph{Shoes}, we use GIST features and color histograms provided by Parikh and Grauman~\cite{ParGra11} as input features to construct the initial rankings $\mbf^I$. For \emph{OSR}, we construct $\mbf^I$ with GIST features from the authors of \cite{KovParGra12}. 

For each target attribute, the remaining attributes are used as the source of test time features. However, our preliminary experiments showed that all target output attributes are strongly correlated, and so applying TUPI with other target attributes as test time features leads to almost perfect results. As such, instead of directly using these ground-truth attributes, we use the outputs of the corresponding individually trained rankers. This corresponds to a practical application scenario where the estimated rankers are denoised by using the other rank estimates as test time information.

\subsection{Results with absolute accuracies and statistical significance tests}
\label{s:results}
\begin{figure*}[tb]
\centering
\includegraphics[width=\linewidth]{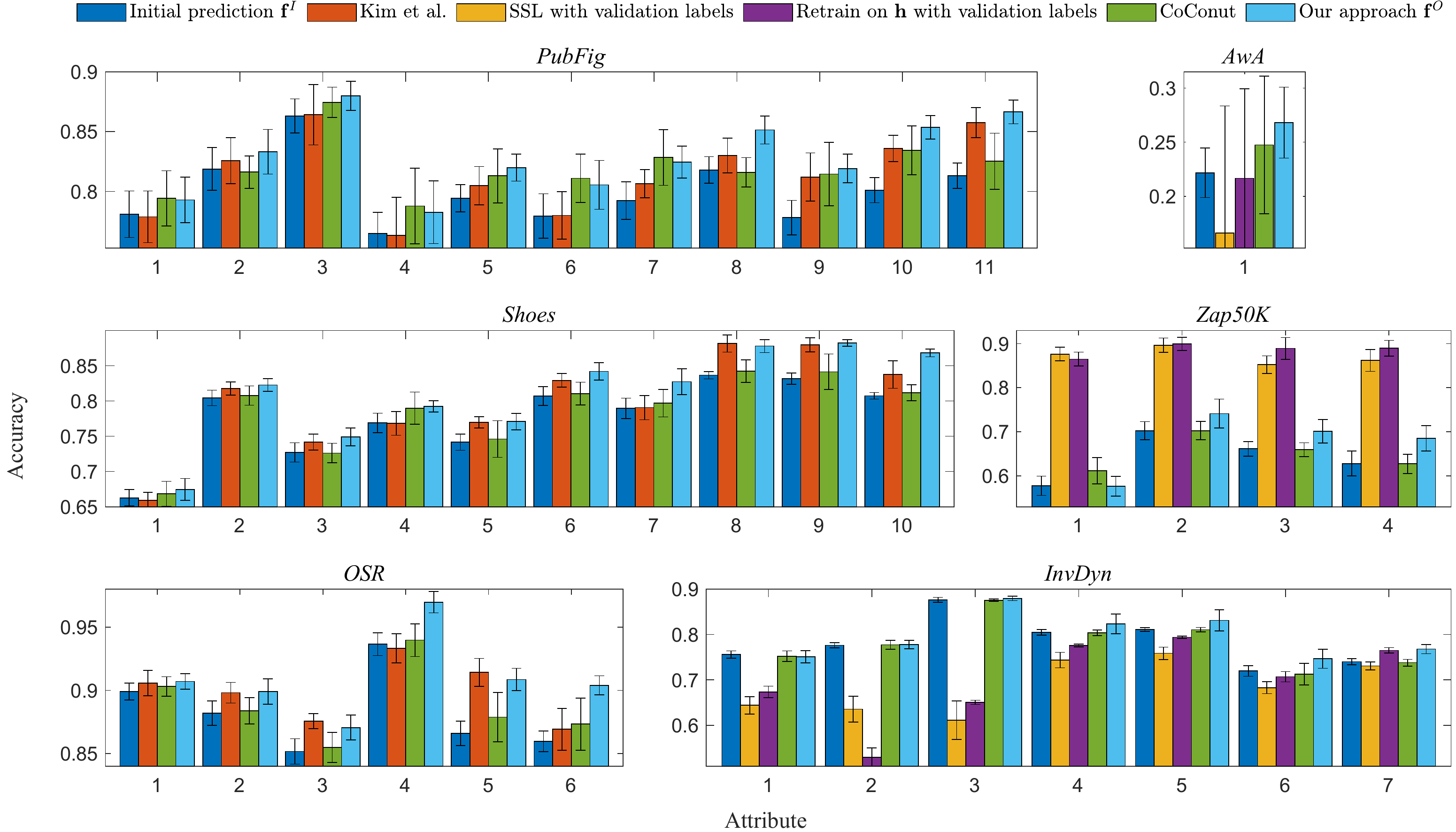}
\caption{Absolute mean accuracies for \emph{PubFig}, \emph{AwA}, \emph{Shoes}, \emph{Zap50K}, \emph{OSR}, and \emph{InvDyn} datasets (higher is better; error bars are standard deviations). The main paper shows performance relative to $\mbf^I$ for easier viewing.}
\label{f:resultsabsolute}
\end{figure*}

In the main paper, we show results relative to $f^I$ for easier interpretation of the bar charts. Figure~\ref{f:resultsabsolute} shows the absolute accuracy results. Further, Table~\ref{t:sigtestresults} shows tests of statistical significance of result differences of different algorithms: 
\begin{itemize}
    \item Our algorithm is better than initial predictions $\mbf^I$ in 87.18\% of cases, and it is not worse in any cases. This shows the effectiveness of exploiting test time features.
    \item Our algorithm is statistically significantly better than Kim~\etal's algorithm in 62.96\% of cases, and it is not worse in any cases. 
    \item Our algorithm is statistically significantly better than SSL in 66.67\% of cases and worse in 33.33\% of cases. The worse cases occurred only on \emph{Zap50K}, where test time features are powerful enough for SSL to be better.
    \item Our algorithm is statistically significantly better than retraining on test time features in 50\% of cases and worse in 33\% of cases. Again, worse cases occurred only on \emph{Zap50K}, where test time features are powerful enough for SSL to be better.
    \item In comparison with CoConut, our algorithm is better in 64.10\% of cases and worse in 2.56\% of cases.
\end{itemize}

Overall, our approach is more useful as an approach in our setting as it provides better performance on average.

\subsection{A simple failure case}
\label{s:simplefailurecase}
We assume that the additional features $\{H^i\}$ are useful if they exhibit strong statistical dependence to the underlying ground-truth predictor $\mbf^*$. However, since we do not have access to $\mbf^*$, we instead measure and enforce statistical dependence to the main predictor $\mbf(t)$ that is being denoised through iterative optimization (Algorithm~\ref{a:mainalgsup} and Eq.~\ref{e:denoisingalgorithmsup}). While our experiments on real-world problems have demonstrated the effectiveness of this approach, simple failure cases exist. For example, if all additional features are identical to the initial predictions, trivially there is no gain. Table~\ref{t:mfeatfailurecase} shows slightly more involved cases: Here, on the \emph{MFeat} dataset, we gradually increase the number of copies of initial predictors $\mbf^I$ in the original reference sets (each containing 5 features). When there is only a single copy of $\mbf^I$ included in the reference set, for all features except for F4, the performance is roughly on par with the case where the original references are used ($\mbf^{O}$). However, as the number of $\mbf^I$ copies increases, the accuracy decreased rather rapidly. 

\begin{table}[t]
\caption{\textbf{A simple failure case on the \emph{MFeat} dataset.} Ranking algorithm mean accuracy percent, plus standard deviation in parenthesis, given the F1--F6 features.\medskip\newline
$\mbf^I$: The initial predictions.\newline
$\mbf^{O}$: TUPI with other F-feature sets as test time information.\newline
$\mbf^{F_i}$: $\mbf^{O}$ with other F-feature sets plus $i$ copies of $\mbf^I$ as test time information. Accuracy decreases as more copies are added.
}
\label{t:mfeatfailurecase}
\vspace{0.25cm}
\resizebox{\linewidth}{!}
{
\begin{tabular}{l c c c c c c}
\toprule
 & F1 & F2 & F3 & F4 & F5 & F6\\
\midrule
% Baseline
\multirow{2}{*}{$\mbf^I$} &77.85  &79.28  &75.70  &70.88  &76.05  &77.10  \\
&(2.26)  &(1.23)  &(2.38)  &(1.23)  &(2.66)  &(1.60) \\
\midrule
% Ours
\multirow{2}{*}{$\mbf^O$} &81.97  &81.45  &78.31  &74.33  &78.19  &82.25  \\
 &(2.95)  &(1.69)  &(3.13)  &(5.00)  &(3.29)  &(2.09)  \\
\midrule
% Gradually adding $\mbf$
\multirow{2}{*}{$\mbf^{F_1}$}  & 82.81& 81.92& 79.22& 71.86& 79.65&82.44\\
  &(2.35) &(1.84), &(2.95) &(2.54) &(3.20) &(1.90) \\
\multirow{2}{*}{$\mbf^{F_3}$}  & 80.50& 81.34& 77.93& 71.12& 78.20&79.83\\
  &(2.16) &(1.43) &(2.75) &(1.64) &(2.83) &(1.65) \\
\multirow{2}{*}{$\mbf^{F_5}$}  & 79.54& 80.80& 77.17& 70.95& 77.54&78.76\\
  &(2.23) &(1.34)  &(2.54) &(1.47) &(2.76) &(1.53) \\
\multirow{2}{*}{$\mbf^{F_7}$}  & 79.24& 80.48& 76.88& 70.92& 77.30&78.37\\
  &(2.23) &(1.29) &(2.47) &(1.48) &(2.71) &(1.49) \\
\multirow{2}{*}{$\mbf^{F_9}$}  &79.06 & 80.31& 76.73& 70.90& 77.15&78.16\\
  &(2.23) &(1.27) &(2.43) &(1.46) &(2.79) &(1.47)  \\
\multirow{2}{*}{$\mbf^{F_{11}}$}  & 78.96& 80.19& 76.65& 70.89& 77.07&78.05\\
  &(2.23) &(1.25) &(2.41) &(1.46) &(2.75) &(1.47) \\
\bottomrule
\end{tabular}
}
\end{table}

%%%%%%%%%%%%%%%%%%%%%%%%%%%%%%%%%%%%%%%%%%%%%%%
\section{TUPI adaptation of Kim~\etal's algorithm and CoConut}
\label{s:adaptation}
\subsection{Adapting Kim~\etal~\cite{KimTomRic17}}
\label{s:kimadaptation}
This method forms predictive \emph{distributions} from reference tasks, then penalizes their pairwise Kullback-Leibler divergence from the target distribution. This implies metric comparison: If a test time feature is the negative of the perfect prediction, then the KL-divergence will be large and Kim~\etal would penalize it; however, the feature is \emph{still statistically dependent}, and our approach would exploit this.

To adapt their method to our setting, if we let their reference task predictions be new features $\{H^i\}$, then this approach works when the (probability) space of each feature coincides with the space of predictions $\calY$. This makes their algorithm applicable only to datasets with one-dimensional test time features as the predictions made for potentially-related tasks (\eg, the \emph{PubFig}, \emph{Shoes} and, \emph{OSR} datasets in the experiments). Figure~\ref{f:resultsabsolute} demonstrates that our general multi-dimensional test time feature algorithm is a strong alternative to Kim~\etal's approach even in this special setting.

\subsection{Adapting CoConut~\cite{KhaLam15}}
\label{s:coconut}

The CoConut framework was developed for \emph{co-classification} problems where multiple data instances are jointly classified. The authors propose to apply a graph Laplacian-type regularizer via adopting the \emph{Cluster Assumption}~\cite{Lux07} to improve during testing the classifications once predicted. In general, a graph Laplacian-type regularizer is defined based on the pairwise similarities of predictions weighted by the corresponding input feature similarities. In CoConut, the new regularizer can also be constructed based on additional features that are available at test time, facilitating a TUPI-like scenario. Please note that we adopt the  mathematical notations from our main paper which differ from the notation of the original CoConut paper~\cite{KhaLam15}. 

Suppose that we have a set of test data features $G=\{\mbg_1,\ldots,\mbg_n\}\subset \calG$ and our goal is to predict a classification label vector $\mbf=[f_1,\ldots,f_n]^\top$ where the value of each element $f_i$ is assigned from a label set $\calY=\{1,\ldots,L\}$. Further, we assume that a set of \emph{base classifiers} $\{(f^I)^l:\calG\to\R\}_{l=1}^L$ are constructed such that $(f^I)^l(\mbg)$ provides a \emph{confidence} that the sample point $\mbg$ belongs to class $l$. Based only on the base classifiers, the initial class label prediction $y^I_i$ for the $i$-th test data point can be made as
\begin{align}
\label{e:youtput}
y^I_i=\argmax_{l=1,\ldots,L}(f^I)^l(\mbg_i).
\end{align}
CoConut improves this \emph{initial predictions} $\mby^I=[y_1,\ldots,y_n]^\top$ by minimizing the following energy\footnote{In $\calO'$, the initial predictions $\mby^I$ are only indirectly taken account via $\{f_l(\mbg_i)\}$.}
\begin{align}
\label{e:orgcoconutsup}
\calO'(\mbf) = &-\sum_{i=1}^n\sum_{l=1}^L \mathbbm{1}[f_i=l](f^I)^l(\mbg_i)\nonumber\\
&+\lambda^C \sum_{i=1}^n\frac{1}{|N_i|}\sum_{\mbg_j\in N_i}w_{ij}\mathbbm{1}[f_i\neq f_j],
\end{align}
where $\mathbbm{1}[\cdot]$ is the indicator function, $\lambda^C\geq 0$ is the regularization hyperparameter, and $N_i$ is the neighbors of $\mbg_i$ in $\calG$ that Khamis and Lampert~\cite{KhaLam15} defined as the $k$-nearest neighbors (NNs). The first term ensures that the final solution does not deviate significantly from the initial class assignments $\mby^I$, while the second term enforces smoothness in the final solution measured in the pairwise similarities of output values weighted by $\{w_{ij}\}$. The weight $w_{ij}$ is defined based on the pairwise similarity of the input features $\mbg_i$ and $\mbg_j$:
\begin{align}
\label{e:weight}
w^C_{ij}=\exp\left(-\frac{d^2_\calG(\mbg_i-\mbg_j)}{(\sigma^C)^2}\right),
\end{align}
where $d^2_\calG$ is a distance measure on $\calG$ and $\sigma^C$ is a hyperparameter. When an additional set of \emph{test time} features $H=\{\mbh_1,\ldots,\mbh_n\}\subset \calH$ is provided, $d^2_\calG$ can be replaced by $d^2_\calG + d^2_\calH$ taking both features into account. This corresponds to a TUPI-like usage of test time features. The discrete optimization problem of minimizing $\calO'$ can be approximately solved based on convex relations. The authors proposed to tune hyperparameter $\lambda^C$ based on the training set used for building the base classifiers $\{(f^I)^l\}_{l=1}^L$. This requires the training labels in testing; we will discuss this in the `tuning hyperparameters' paragraph of the next subsection.

\paragraph{CoConut adaptation applied to Relative Attributes ranking.} 

As the original CoConut optimization problem (Eq.~\ref{e:orgcoconutsup}) was designed for discrete classification problems, it needs to be \emph{adapted} before it can be applied to Relative Attributes (RA) ranking problems where the outputs of the base predictors take continuous values. First, it should be noted that when $\{f_i\}$ and $\{(f^I)^l\}$ take continuous values, counting the occurrence of equal values via the indicator evaluations ($\mathbbm{1}[\cdot]$) in Eq.~\ref{e:orgcoconutsup} leads to zero values in the first term of Eq.~\ref{e:orgcoconutsup} in general. Instead, we reinterpret the first term as the measure of deviation (per test instance) between the hypothesized solution $f_i$ and the initial base prediction $f^I_i$: $f^I_i:=\max_{l=1,\ldots,L}(f^I)^l(\mbg_i)$ is the confidence of predicting the label $y^I_i$ for $\mbg_i$~\cite{KhaLam15} in the original classification setting (see Eq.~\ref{e:youtput}). Instantiating this interpretation in the real-valued prediction case, we cast the first term in $\calO'$ into\hspace{-0.5cm}
\begin{align}
\calO_1'(\mbf) = \sum_{i=1}^n (f^I_i-f_i)^2
\end{align}
measuring the deviation between $\mbf=[f_1,\ldots,f_n]^\top$ and $\mbf^I=[f^I_1,\ldots,f^I_n]^\top$. Unlike the ($L$-class) classification problems initially considered by Khamis and Lampert~\cite{KhaLam15}, we do not have $L$ different base predictors (one per class). Therefore, a single base predictor $\mbf^I$ is used.

Now, relaxing the equality constraints in the second regularization term of $\calO'$ into a measure of continuous squared deviations, and adopting the $k$-NN structure for $\{N_i\}$ (with $k^C$ neighbors) as used in the original CoConut setting, the second term can be restated as 
\begin{align}
\calO_2'(\mbf) = \frac{\lambda^C}{k^C} \mbf^\top L \mbf,
\end{align}
where $L$ is the graph Laplacian as $L=D^C-W^C$, and
\begin{align}
W^C_{ij} = \begin{cases} 
      w^C_{ij} & \text{ if } \mbg_j\in N_i \text{ (see Eq.~}\ref{e:weight}) \\
      0 & \text{otherwise}.
   \end{cases}
\end{align}
and $[D^C]$ is a diagonal matrix of row sums of $W^C$: $[D^C]_{ii}=\sum_j [W^C]_{ij}$. $\calO_1'$ ensures that the final solution does not deviate significantly from $\mbf^I$, and $\calO_2'$ contributes to improving the final solution by enforcing its spatial smoothness measured via the Laplacian $L$.

\paragraph{Tuning hyperparameters.}
In the original CoConut setting, the authors proposed to tune the hyperparameter $\lambda_C$ (Eq.~\ref{e:orgcoconutsup}) based on performance on the training set that is used to train the base predictors $\{(f^I)^l\}$. This requires access to training labels at test time. However, in our TUPI scenario, a large set of labeled training data points is not available: If such training labels are available, a better alternative to TUPI is often simply to re-train the baseline predictor $f^I$ with the original $G$ and the test time features $H$. Indeed, in our preliminary RA experiments using deep neural networks as baselines, this constantly led to better performance than our TUPI algorithm as well as our CoConut adaptation. 

Furthermore, we also observed in preliminary experiments that for the problem of estimating continuous predictions in the RA setting, tuning the CoConut hyperparameters in this way suffered from overfitting and it led to worse results than the original predictors $f^I$. 

Therefore, in our experiments provided in the main paper, we tune the hyperparameters based on separate validation sets. The hyperparameters include $k^C$ for $k$-NNs, $\sigma^C$ (Eq.~\ref{e:weight}), and $\lambda^C$. We adaptively decided $\sigma^C$ as twice the mean distance within each $N_i$ as suggested by Hein and Maier~\cite{HeiMai07}. The other two parameters $k^C$ and $\lambda^C$ are tuned based on the validation accuracies. In the original CoConut framework, the authors tuned only one parameter $\lambda^C$ and the selection of $k^C$ was not discussed. We observed that the best choice of $k^C$ differs across different datasets, and the impact of varying this value on the final results is substantial. Thus, we concluded that $k^C$ also needs to be tuned per dataset.

\paragraph{CoConut adaptation summary.} We minimize the energy\hspace{-0.5cm}
\begin{align}
\label{e:adaptcoconut}
\calO'(\mbf) = \|\mbf-\mbf^I\|^2+\frac{\lambda^C}{k^C} \mbf^\top L \mbf,
\end{align}
where $L$ is the graph Laplacian calculated based on local $k$-nearest neighbors (with $k=k^C$) in the test time feature space. The first term in $\calO'$ ensures that the final solution does not deviate significantly from $\mbf^I$ while the second term contributes to improving the final solution by enforcing its spatial smoothness measured via the Laplacian $L$. The two hyperparameters $\lambda^C$ and $k^C$ are tuned based on validation accuracies similarly to our algorithm.

%%%%%%%%%%%%%%%%%%%%%%%%%%%%%%%%%%%%%%%%%%%%%%%
\begin{figure*}[t]
\centering
\includegraphics[width=\linewidth]{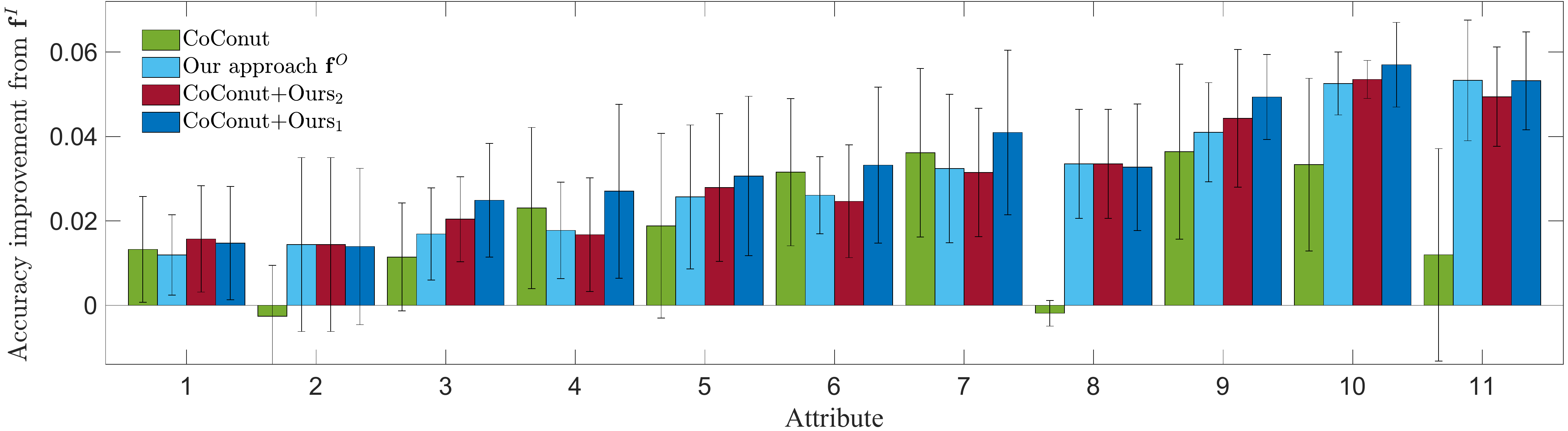}
\caption{Accuracy improvements over $\mbf^I$ for \emph{PubFig} dataset where CoConut and our algorithm demonstrate complementary strengths.
}
\label{f:TUPICoConutComb}
\end{figure*}

\subsection{Combining CoConut and our algorithm}
\label{s:coconuttupicombination}

We observed in the experiments that, overall, our algorithm provides higher accuracy than CoConut (Fig.~\ref{f:resultsabsolute}). At the same time, the specific results on the \emph{PubFig} dataset demonstrate that CoConut and our algorithm have complementary strengths: Our algorithm generates the best results in attributes 2, 3, 5, and 8--11 while CoConut is the best on attributes 1, 4, 6, and 7. For these attributes, we observe noticeable accuracy differences in the corresponding results of CoConut and our algorithm. 

As such, we developed two new algorithms which combine the benefits of CoConut and our algorithm. Our first combination attempt `CoConut+Ours$_1$' algorithmically combines the two. This algorithm minimizes a new energy $\calO''$ which combines the energy functional of our algorithm with the graph Laplacian regularizer of CoConut:
\begin{align}
\calO''(\mbf;t) &= \calO(\mbf;t)+\calO_2'(\mbf).
\label{e:tupicoconutcomb}
\end{align}
5
The resulting new algorithm leverages the global statistical dependence present among multiple features (via our denoising strategy) as well as the local spatial smoothness of the predictor variables (via the Coconut regularizer). Figure~\ref{f:TUPICoConutComb} shows the results: Indeed, combining these benefits, CoConut+Ours$_1$ often generates the best results (attributes 3--7, 9--11). More importantly, even when it is not the best, the corresponding accuracy is close to the best except for the third attribute where our original algorithm is clearly better. However, a drawback of this approach is that to obtain these results, it required tuning four hyperparameters ($\lambda$ and $\sigma_w^2$ from our algorithm and $\lambda^C$ and $k^C$ from CoConut), which is often prohibitively expensive in practical applications.

Our second algorithm is computationally affordable: CoConut+Ours$_2$ \emph{selects} either of the outputs of CoConut and our algorithm based the validation accuracy: For both algorithms, the predictions are independently generated and these predictions with higher validation accuracy are selected per prediction set as the final outputs. Figure~\ref{f:TUPICoConutComb} demonstrates that while this approach is overall worse than CoConut+Ours$_1$ and often even worse than either of the CoConut or our algorithm, it still delivers performance that does not deviate significantly from the best results among the CoConut and our algorithm per attribute. This shows that CoConut+Ours$_2$ facilitates trading the hyperparameter tuning complexity of CoConut+Ours$_1$ with the final prediction accuracies.

%%%%%%%%%%%%%%%%%%%%%%%%%%%%%%%%%%%%%%%%%%%%%%%
\section{Additional future work}
\label{s:addfuturework}
In our application scenario, we assumed no access to the underlying ranking function $f$ and focused on evaluating $f^O$ on a fixed set of points $G$. When an explicit functional form of $f^O$ is required, \eg when one wishes to apply $f^O$ to new test points $\mbg_\text{new}\notin G$, two scenarios are possible: 1) If we remove the assumption that the parametric form of $f$ is unknown, one could apply our algorithm to tune the parameter vector $\mbw$ of $f_\mbw$. Since our objective function $\calO$ (Eq.~\ref{e:denoisingalgorithmsup}) is smooth, this approach is straightforward when $f_\mbw$ is continuously differentiable with respect to $\mbw$ (which is the case for DNNs, RSVMs, and many other predictors); 2) If $f$ form remains unavailable, one could train a smooth regressor $f^O$ on the large set of inputs $G$ using the corresponding non-parametric estimates $\mbf^O$ as labels.

\begin{table*}[p]
\caption{Results of t-test with $\alpha=0.95$ for relative accuracy differences of different algorithms. 1 and -1: statistically significantly positive and negative, respectively and 0: statistically insignificant.\medskip\newline
$\mbf^I$: \hspace{0.9mm}The initial predictions.\newline
$\mbf^{K}$: Kim~\etal's algorithm~\cite{KimTomRic17}.\newline
$\mbf^{S}$: \hspace{0.8mm}Semi-supervised learning on $\mbh$.\newline
$\mbf^{R}$: \hspace{0.4mm}Retrain on $\mbh$.\newline
$\mbf^{C}$: \hspace{0.4mm}CoConut~\cite{KhaLam15}.\newline
$\mbf^{O}$: \hspace{0.4mm}Our TUPI algorithm.\newline
}
\label{t:sigtestresults}
\centering
\resizebox{0.8\linewidth}{!}{
\begin{tabular}{ccccccccccc}
\toprule
Dataset &Attr. &$\mbf^{K}-\mbf^I$ & $\mbf^{S}-\mbf^I$ & $\mbf^{R}-\mbf^I$ & $\mbf^{C}-\mbf^I$ & $\mbf^{O}-\mbf^I$ &$\mbf^{O}-\mbf^{K}$ &$\mbf^{O}-\mbf^{S}$ & $\mbf^{O}-\mbf^{R}$ & $\mbf^{O}-\mbf^{C}$\\
\midrule
\multirow{11}{*}{\emph{Pubfig}}
& 1 & 0 & N/A & N/A & 1 & 1 & 1 & N/A & N/A & 0\\
& 2 & 0 & N/A & N/A & 0 & 0 & 0 & N/A & N/A & 1\\
& 3 & 0 & N/A & N/A & 1 & 1 & 1 & N/A & N/A & 0\\
& 4 & 0 & N/A & N/A & 1 & 1 & 1 & N/A & N/A & 0\\
& 5 & 0 & N/A & N/A & 1 & 1 & 1 & N/A & N/A & 0\\
& 6 & 0 & N/A & N/A & 1 & 1 & 1 & N/A & N/A & 0\\
& 7 & 1 & N/A & N/A & 1 & 1 & 1 & N/A & N/A & 0\\
& 8 & 1 & N/A & N/A & 0 & 1 & 1 & N/A & N/A & 1\\
& 9 & 1 & N/A & N/A & 1 & 1 & 0 & N/A & N/A & 0\\
& 10 & 1 & N/A & N/A & 1 & 1 & 1 & N/A & N/A & 1\\
& 11 & 1 & N/A & N/A & 0 & 1 & 1 & N/A & N/A & 1\\
\midrule
\multirow{7}{*}{\emph{InvDyn}}
& 1 & N/A & -1 & -1 & 0 & 0 & N/A & 1 & 1 & 0\\
& 2 & N/A & -1 & -1 & 0 & 0 & N/A & 1 & 1 & 0\\
& 3 & N/A & -1 & -1 & 0 & 0 & N/A & 1 & 1 & 1\\
& 4 & N/A & -1 & -1 & 0 & 1 & N/A & 1 & 1 & 1\\
& 5 & N/A & -1 & -1 & 0 & 1 & N/A & 1 & 1 & 1\\
& 6 & N/A & -1 & 0 & 0 & 1 & N/A & 1 & 1 & 1\\
& 7 & N/A & -1 & 1 & 0 & 1 & N/A & 1 & 0 & 1\\
\midrule
\multirow{1}{*}{\emph{AwA}}
& 1 & N/A & 0 & 0 & 0 & 1 & N/A & 1 & 0 & 0\\
\midrule
\multirow{10}{*}{\emph{Shoes}}
& 1 & 0 & N/A & N/A & 0 & 1 & 1 & N/A & N/A & 0\\
& 2 & 1 & N/A & N/A & 0 & 1 & 0 & N/A & N/A & 1\\
& 3 & 1 & N/A & N/A & 0 & 1 & 1 & N/A & N/A & 1\\
& 4 & 0 & N/A & N/A & 1 & 1 & 1 & N/A & N/A & 0\\
& 5 & 1 & N/A & N/A & 0 & 1 & 0 & N/A & N/A & 1\\
& 6 & 1 & N/A & N/A & 0 & 1 & 1 & N/A & N/A & 1\\
& 7 & 0 & N/A & N/A & 0 & 1 & 1 & N/A & N/A & 1\\
& 8 & 1 & N/A & N/A & 0 & 1 & 0 & N/A & N/A & 1\\
& 9 & 1 & N/A & N/A & 0 & 1 & 0 & N/A & N/A & 1\\
& 10 & 1 & N/A & N/A & 0 & 1 & 1 & N/A & N/A & 1\\
\midrule
\multirow{6}{*}{\emph{OSR}}
& 1 & 0 & N/A & N/A & 0 & 1 & 0 & N/A & N/A & 0\\
& 2 & 1 & N/A & N/A & 0 & 1 & 0 & N/A & N/A & 1\\
& 3 & 1 & N/A & N/A & 0 & 1 & 0 & N/A & N/A & 1\\
& 4 & 0 & N/A & N/A & 0 & 1 & 1 & N/A & N/A & 1\\
& 5 & 1 & N/A & N/A & 0 & 1 & 0 & N/A & N/A & 1\\
& 6 & 0 & N/A & N/A & 1 & 1 & 1 & N/A & N/A & 1\\
\midrule
\multirow{4}{*}{\emph{Zap50K}}
& 1 & N/A & 1 & 1 & 1 & 0 & N/A & -1 & -1 & -1\\
& 2 & N/A & 1 & 1 & 0 & 1 & N/A & -1 & -1 & 1\\
& 3 & N/A & 1 & 1 & 0 & 1 & N/A & -1 & -1 & 1\\
& 4 & N/A & 1 & 1 & 0 & 1 & N/A & -1 & -1 & 1\\
\bottomrule
\end{tabular}
}
\end{table*}

\end{document}

%% file: kimkimacros.tex
\usepackage{mathtools}
\usepackage{xspace}

\newcommand{\mbf}{\mathbf{f}}
\newcommand{\mbg}{\mathbf{g}}
\newcommand{\mbh}{\mathbf{h}}

\newcommand{\mbp}{\mathbf{p}}
\newcommand{\mbq}{\mathbf{q}}

\newcommand{\mbv}{\mathbf{v}}
\newcommand{\mbw}{\mathbf{w}}
\newcommand{\mbx}{\mathbf{x}}
\newcommand{\mby}{\mathbf{y}}

\newcommand{\mbC}{\mathbf{C}}

\newcommand{\mbK}{\mathbf{K}}

\newcommand{\mbS}{\mathbf{S}}

\newcommand{\E}{\mathbb{E}}

\newcommand{\mbbP}{\mathbb{P}}
\newcommand{\R}{\mathbb{R}}

\newcommand{\calC}{\mathcal{C}}

\newcommand{\calG}{\mathcal{G}}
\newcommand{\calH}{\mathcal{H}}

\newcommand{\calK}{\mathcal{K}}

\newcommand{\calM}{\mathcal{M}}

\newcommand{\calO}{\mathcal{O}}

\newcommand{\calV}{\mathcal{V}}
\newcommand{\calW}{\mathcal{W}}
\newcommand{\calX}{\mathcal{X}}
\newcommand{\calY}{\mathcal{Y}}

\def\R{\mathbb{R}}

\def\calX{\mathcal{X}}

\def\tr{\mathop{\rm tr}\nolimits}

\def\argmax{\mathop{\rm arg\,max}\limits}%    a math operator.
\def\ones{\mathbf{1}}
\newcommand{\ignore}[1]{}

\makeatletter
\DeclareRobustCommand\onedot{\futurelet\@let@token\@onedot}
\def\@onedot{\ifx\@let@token.\else.\null\fi\xspace}
\def\eg{{e.g}\onedot}

\def\ie{{i.e}\onedot}

\def\cf{{cf}\onedot}

\def\etal{{et al}\onedot}
\makeatother